\documentclass[letterpaper]{article}
\usepackage{aaai2026} 
\usepackage{times} 
\usepackage{helvet} 
\usepackage{courier} 
\usepackage[hyphens]{url} 
\usepackage{graphicx} 
\usepackage{graphicx}  
\usepackage{natbib}  
\usepackage{caption}  
\usepackage[ruled,vlined]{algorithm2e}

\usepackage{newfloat}
\usepackage{listings}
\DeclareCaptionStyle{ruled}{labelfont=normalfont,labelsep=colon,strut=off} 
\usepackage{amsfonts}       
\usepackage{nicefrac}       
\usepackage{amsthm}
\usepackage{amsmath}
\usepackage{mathtools}
\usepackage{subcaption}

\newcommand{\presec}{\vspace{0.00in}}
\newcommand{\postsec}{\vspace{0.0in}}

\newtheorem{lemma}{Lemma}
\newtheorem*{lemma*}{Lemma}

\newtheorem*{theorem*}{Theorem}
\theoremstyle{definition}

\newtheorem*{definition*}{Definition}

\newcommand{\aname}{KVReviver}

\title{KVReviver: Reversible KV Cache Compression with Sketch-Based Token Reconstruction}

\author {
Aomufei Yuan\textsuperscript{\rm1}\thanks{Preprint.\\This work was done during an internship in ByteDance Inc.},
Zhiming Wang\textsuperscript{\rm1},
Ruijie Miao\textsuperscript{\rm1},
Dayu Wang\textsuperscript{\rm1},
Yuxuan Tian\textsuperscript{\rm1},
Zihan Wang\textsuperscript{\rm1},
Yebo Peng\textsuperscript{\rm1},
Yuhan Wu\textsuperscript{\rm1},
Bairen Yi\textsuperscript{\rm2},
Xin Liu\textsuperscript{\rm2},
Tong Yang\textsuperscript{\rm1},
}
\affiliations{
    \textsuperscript{\rm 1}Peking University\\
    \textsuperscript{\rm 2}ByteDance Inc.\\

    YuanAomufei@gmail.com
}

\begin{document}

\maketitle

\begin{abstract}

As the context length of current large language models (LLMs) rapidly increases, the memory demand for the Key-Value (KV) cache is becoming a bottleneck for LLM deployment and batch processing. Traditional KV cache compression methods typically involve permanently evicting or irreversibly merging "less important" tokens with low attention scores. This approach results in the unrecoverable loss of token information, which we call \textbf{Contextual Amnesia}, significantly degrading the model's information retrieval capability. To address this issue, we propose \aname{}, a reversible KV cache compression method based on the sketch algorithm. This method allows reconstructing compressed tokens from an additional data structure, thus enabling full-scale computation within limited memory. Experiments showed that in 2k-length contexts, it requires only 10\% of KV Cache budget while maintaining identical end-to-end inference accuracy. For 32k-length contexts, it achieves equivalent or comparable accuracy (\(\sim\)2\% accuracy loss) using merely 25\% of KV Cache budget.

\end{abstract}

\presec
    \section{Introduction}
    \label{sec:intro}
\postsec

\subsection{Motivation}

Compressing the memory usage of the KV cache at the algorithmic level has become an urgent task. With the rapid development of large language models (LLMs) based on the Transformer architecture \citep{transformer}, models supporting longer context windows are increasingly prevalent, greatly expanding the potential applications of AI. However, in the self-attention mechanism of Transformer models, KV cache mechanism is introduced to store token embeddings that are repeatedly used in order to avoid redundant computations. The problem arises when the context window increases: the memory occupied by the KV cache grows linearly with the window length, eventually becoming the bottleneck for memory usage, surpassing model parameters in long-context scenarios. For example, DeepSeek-R1-0528-Qwen3-8B \citep{deepseekai2025deepseekr1incentivizingreasoningcapability} supports a maximum context length of 128k, whose KV cache consumes around 20GB of GPU memory, while the model parameters only occupy about 16GB. Moreover, when we batch the input sequence, which is extremely common in real inference systems, the size of KV cache multiplies by the batch size, easily exceeding the memory limit of GPUs. 



In existing KV cache compression methods, one important approach is to reduce the number of tokens in the cache to lower memory usage. These methods can generally be divided into two categories: eviction and merging. Eviction-based methods \citep{streamingllm, h2o, snapkv, pyramidinfer, pyramidkv} typically involve selecting a subset of tokens using a specific algorithm and permanently evicting the other tokens. Merging-based methods \citep{cam, dmc, d2o} rearrange and group tokens in a certain way, storing only the ``new token'' generated by the combination of tokens within the same group, thus reducing the number of tokens stored. However, both methods have an inherent problem, which we call \textbf{Contextual Amnesia}: since the evicted or merged tokens cannot be reconstructed, there is irreversible information loss of certain tokens.

\begin{figure*}[h]
\centering
\begin{subfigure}{0.33\textwidth}
    \centering
    \includegraphics[width=\textwidth]{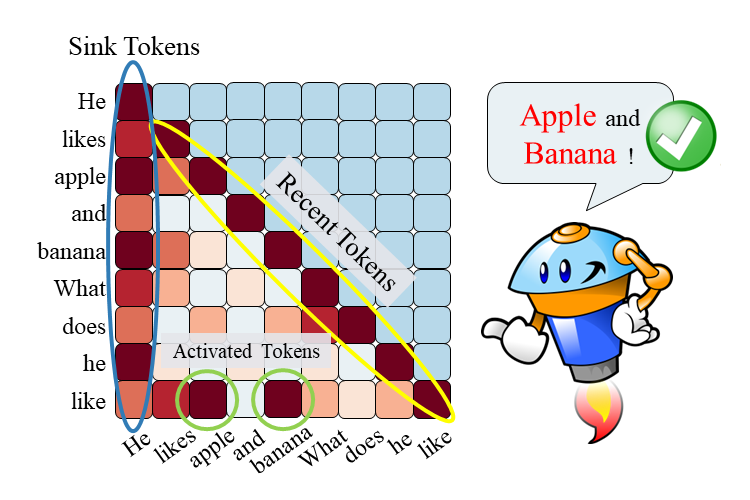}
    \caption{Full KV Cache}
    \label{fig:amnesia:full}
\end{subfigure}
\hspace{-0.2cm}
\begin{subfigure}{0.33\textwidth}
    \centering
    \includegraphics[width=\textwidth]{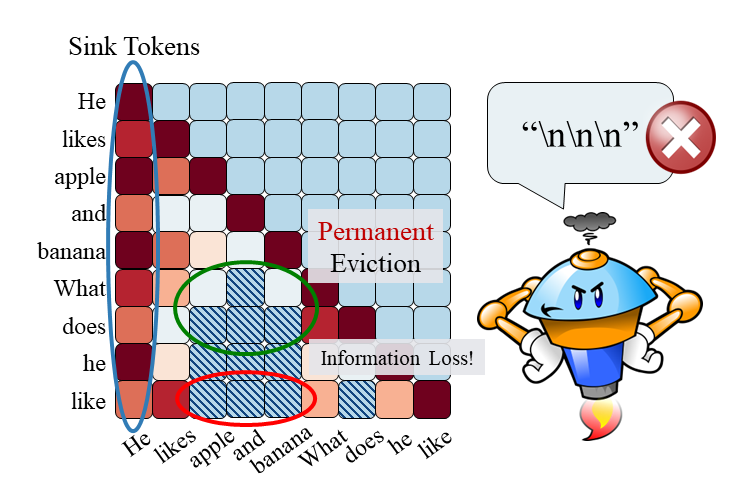}
    \caption{Eviction}
    \label{fig:amnesia:eviction}
\end{subfigure}
\hspace{-0.2cm}
\begin{subfigure}{0.33\textwidth}
    \centering
    \includegraphics[width=\textwidth]{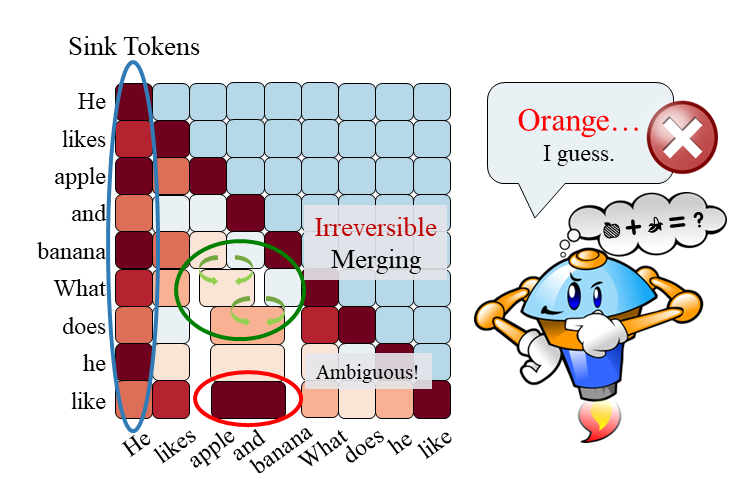}
    \caption{Merging}
    \label{fig:amnesia:merging}
\end{subfigure}
\caption{A token-level view of Contextual Amnesia caused by Eviction and Merging.}
\label{fig:amnesia}
\end{figure*}

Specifically, these methods rely on an assumption, that the information stored in the tokens is sparse, and therefore, using only a subset of the token information can yield similar results to full-scale attention. However, this assumption does not hold for certain downstream tasks. An intuitive example is when the model is provided with an article as background knowledge and interacts with the user's questions based on the content of the article. Some tokens containing crucial information may be overlooked if they were not involved in previous interactions, causing them to be permanently evicted or merged, thus unable to be correctly activated in subsequent questions. Figure \ref{fig:amnesia} shows a simplified example, where the background is ``He likes apple and banana'', and question is ``What does he like''. With full KV cache, the model outputs ``Apple and banana'' normally, but if critical tokens such as ``apple'' and ``banana'' are evicted, it is impossible for the model to fetch the information previously stored within them. And, if ``apple'' and ``banana'' are merged as a new token, the model will be confused by some random ``average'' fruit.



\subsection{Ideas of Design}


To address Contextual Amnesia, we propose \textbf{\aname{}}, a KV cache compression-reconstruction approach that does not rely on the assumption of token sparsity. That is to say, we never consider any token's information as negligible even if it is currently overlooked by subsequent tokens, which is common in practice. With this premise, we have to store the ``less important'' tokens in a more compact manner and maintain its ability to be reconstructed. Sketch, a widely used data structure, fits perfectly in this manner with its high memory utilization efficiency and low information loss when restoring the input. Therefore, \aname{} stores the compressed tokens of the KV cache separately in a sketch, and reconstructs them when we need the KV cache to be involved in self-attention calculation.




\subsection{Contribution}

This paper makes the following contributions:


\begin{itemize}
    \item We are the first to enable the token-level reconstruction of compressed KV cache. We achieve this by integrating the classic sketch data structure with LLMs and designing \aname{}, a novel sketch-based algorithm that demonstrates high compression efficiency and minimal information loss.
    
    \item We study and address the widespread Contextual Amnesia problem caused by traditional KV cache compression methods.

    \item We analyze the mathematical properties of the Key and Value matrices and derive rigorous theoretical foundations based on these properties, ensuring the error bounds of our method.

\end{itemize}




Experimental results show that \aname{} achieves accuracy comparable to full self-attention computation while using only 10\% of the memory on 2k-length contexts, and maintains equivalent or comparable accuracy with approximately 2\% accuracy loss with 25\% KV Cache budget on 32k-length contexts. Such results prove that \aname{} can effectively eliminate Contextual Amnesia caused by traditional compression methods.


\presec
    \section{Background}
    \label{sec:background}

    \subsection{Related Work}

We take LLaMA-2-7B \citep{llama2} as an example to analyze the memory footprint of the KV cache based on the formula in Equation \ref{eq:cachesize}. Various related works have optimized different components of this equation:

\begin{figure*}[t]
    \begin{equation}
    \text{Cache Size} = 
    \underset{\text{\small Key \& Value}}{2} \times
    \underset{\text{\small Layers}}{32} \times
    \underset{\text{\small KV Heads}}{32} \times
    \underset{\text{\small Tokens}}{2048} \times
    \underset{\text{\small Head Dim (Channels)}}{128} \times
    \underset{\text{\small Bytes}}{2} = 1\text{GiB}
    \label{eq:cachesize}
    \end{equation}
\end{figure*}

\begin{figure*}[h]
\centering
\begin{subfigure}{0.49\textwidth}
    \centering
    \includegraphics[width=\textwidth]{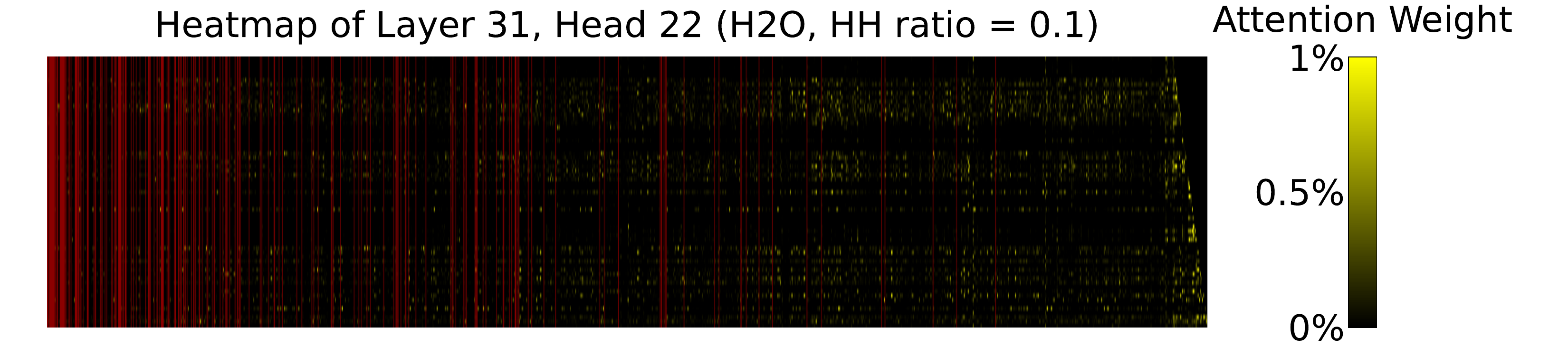}
    \label{fig:amnesia_exp:h2o_0.1_1}
\end{subfigure}
\begin{subfigure}{0.49\textwidth}
    \centering
    \includegraphics[width=\textwidth]{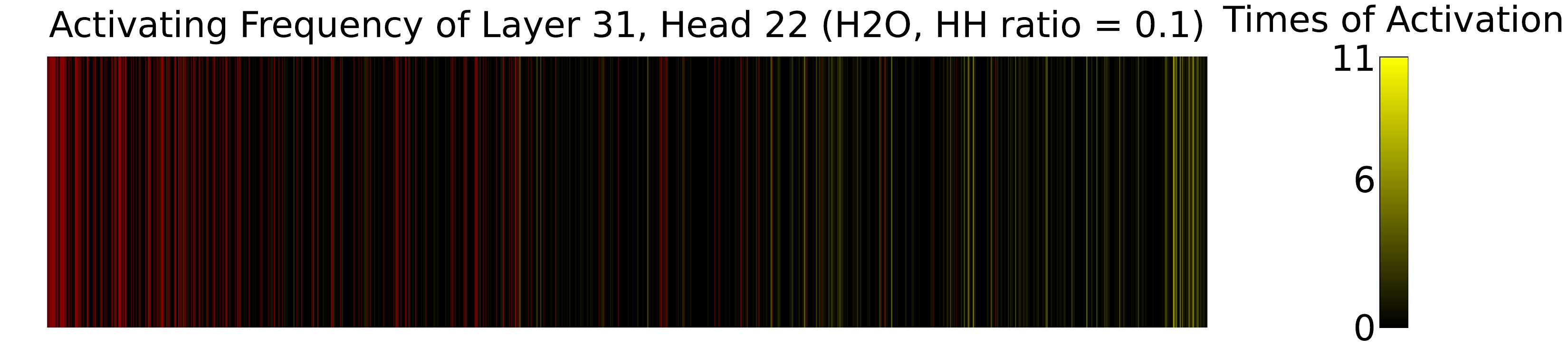}
    \label{fig:amnesia_exp:h2o_0.1_2}
\end{subfigure}
\begin{subfigure}{0.49\textwidth}
    \centering
    \includegraphics[width=\textwidth]{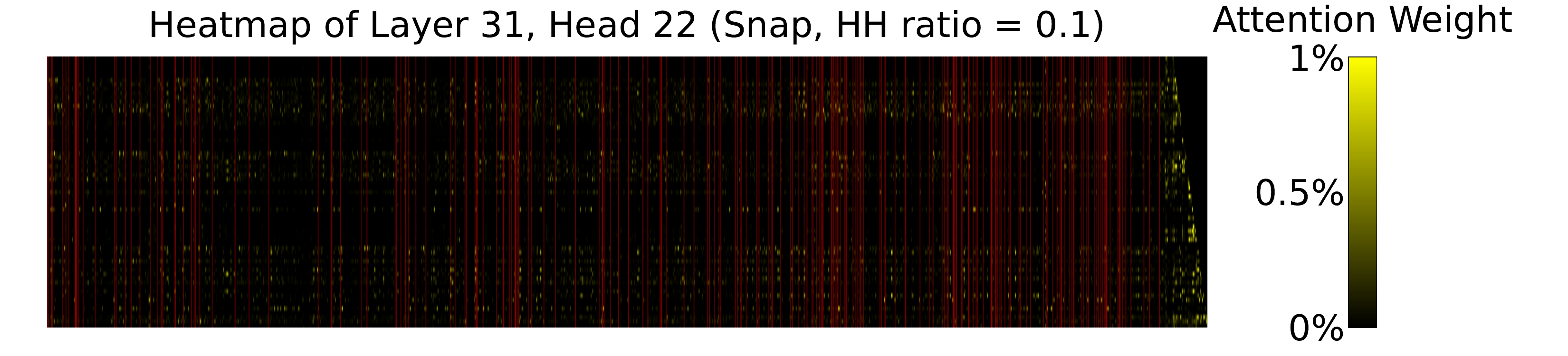}
    \label{fig:amnesia_exp:snap_0.1_1}
\end{subfigure}
\begin{subfigure}{0.49\textwidth}
    \centering
    \includegraphics[width=\textwidth]{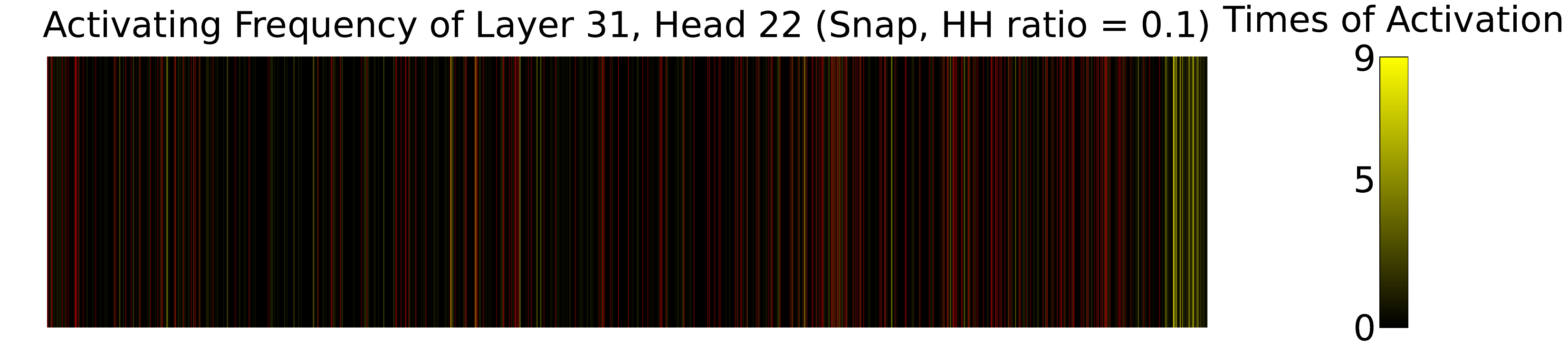}
    \label{fig:amnesia_exp:snap_0.1_2}
\end{subfigure}
\begin{subfigure}{0.49\textwidth}
    \centering
    \includegraphics[width=\textwidth]{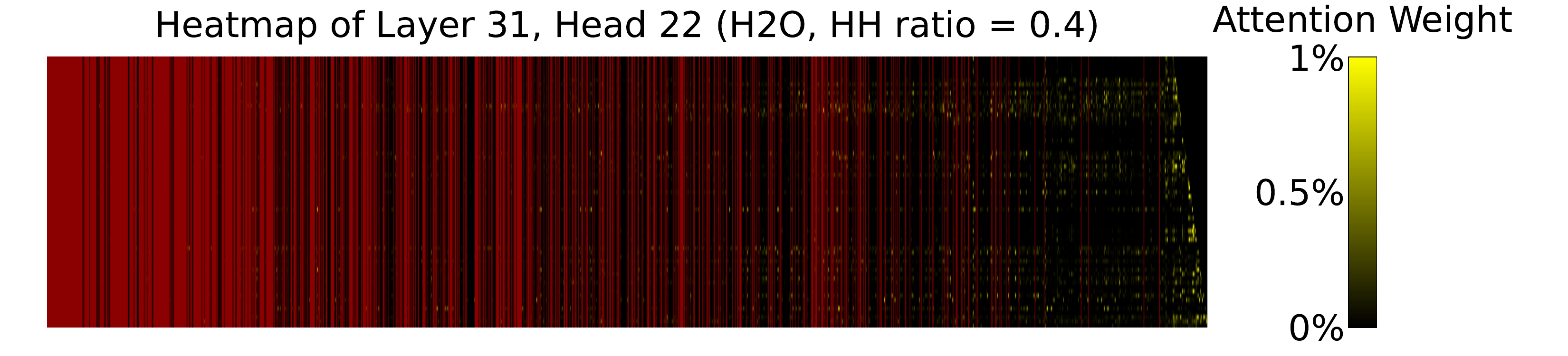}
    \label{fig:amnesia_exp:h2o_0.4_1}
\end{subfigure}
\begin{subfigure}{0.49\textwidth}
    \centering
    \includegraphics[width=\textwidth]{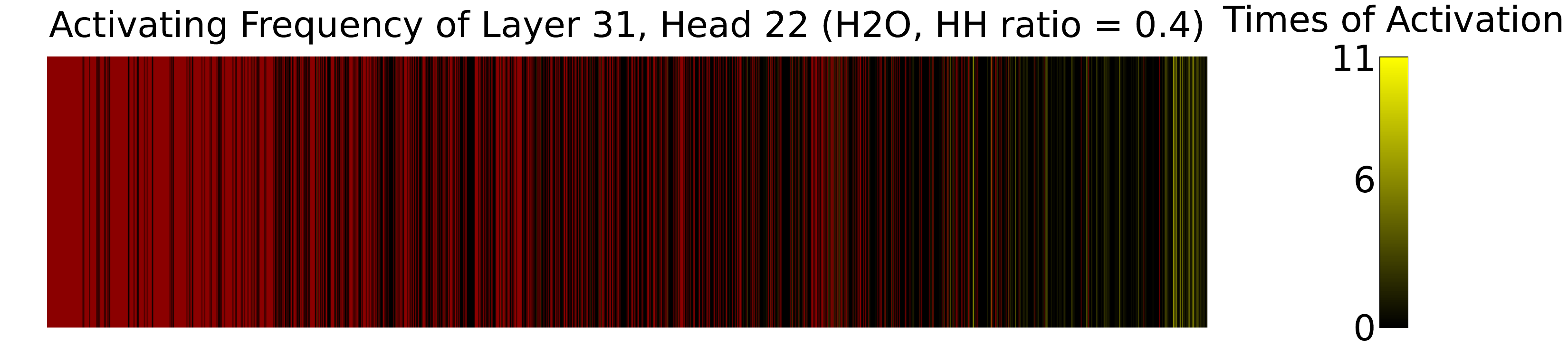}
    \label{fig:amnesia_exp:h2o_0.4_2}
\end{subfigure}
\caption{Heatmap and Activating Frequency of different algorithms and Heavy Hitter ratios.}
\label{fig:amnesia_exp}
\end{figure*}


\begin{itemize}
    \item \textbf {Key \& Value Level}: The DeepSeek MLA architecture \citep{deepseek-v2} projects Key and Value into a shared low-rank representation and later reconstructs them separately in higher dimensions. By storing only the shared latent vector as the KV cache, this approach eliminates the memory overhead factor of 2.


    \item \textbf{Layer Level}: The CLA method \citep{cla} proposes sharing the KV cache across adjacent layers, reducing the number of layers that need to store KV cache. 


    \item \textbf{KV Head Level}: Architectures such as MQA \citep{mqa} and GQA \citep{gqa} reduce the number of KV heads while maintaining the same number of Query heads. During computation, KV heads are replicated for one-to-many operations, allowing storage only of the pre-replication KV cache, reducing the number of heads. Methods like DuoAttention \citep{duoattention} classify heads, storing the full KV cache only in a subset of heads while applying token-level compression to the rest, achieving a similar effect to head reduction.


    \item \textbf{Head Dimension (Channel) Level}: In addition to the MLA architecture \citep{seed-thinking-v1.5}, which applies low-rank compression during pretraining, Palu \citep{palu} employs SVD decomposition on trained Key and Value weight matrices to separate them into low-rank projections and high-dimensional reconstructions. This allows maintaining a KV cache with a reduced head dimension, achieving an MLA-like low-rank effect in standard MHA \citep{transformer} architectures.


    \item \textbf{Byte Level}: KV cache is typically stored with 16-bit floating-point precision. Quantization methods \citep{kvquant, atom} represent KV cache in lower-precision integers (2-bit to 8-bit) during storage and dequantize them back to floating points during computation, reducing per-element memory consumption.


    \item \textbf {Token Level}: This category has been extensively studied and primarily falls into two approaches: eviction and merging.


    \textbf {Eviction-Based Methods}: StreamingLLM \citep{streamingllm} retains only the most recent window of tokens while keeping a few initial tokens, significantly improving inference accuracy. H2O \citep{h2o} accumulates each token’s past attention scores to predict its importance and discards low-importance tokens. Recent works such as SnapKV \citep{snapkv}, PyramidInfer \citep{pyramidinfer}, and PyramidKV \citep{pyramidkv} further optimize budget allocation across layers and refine attention score computation strategies.


    \textbf{Merging-Based Methods}: CaM \citep{cam} merges the Value vectors of tokens with low attention scores into a few retained tokens with a certain probability but does not merge their Key vectors. DMC \citep{dmc} fine-tunes the model to predict whether adjacent tokens should be merged and determine merging weights. D2O \citep{d2o} selects the most similar neighboring Key vectors based on cosine similarity for weighted merging.


\end{itemize}


Notably, except for token-level methods, all other approaches are orthogonal to \aname{}. This allows \aname{} to be combined with any of the aforementioned methods for even higher compression rates. Moreover, token-level approaches inherently suffer from Contextual Amnesia, and we are the first to propose reviving the compressed KV cache to address this issue.

\subsection{Sketch-based Approximation}

\label{sec:background:sketch}

Sketch is an approximate data structure widely used \citep{cmsketch,spacesaving,cocosketch} in fields such as frequency count. Instead of storing raw data directly, Sketch compresses and encodes data, allowing efficient and low-error queries even under extreme memory constraints while maintaining key statistical characteristics such as the unbiased property. Among them, Count Sketch \citep{countsketch} is designed to record data stream frequencies. The sketch employed in this work constitutes a variant of the Count Sketch, whose fundamental mechanism we briefly outline:


Consider a series of consecutive keys. Upon arrival of a key, we need to count the appearance of the same key up to that point. A straightforward approach is to use a hash table (such as a `dict' in Python) to maintain a counter for each key. However, this method requires \(O(m)\) memory, where \(m\) is the number of distinct keys.



The Count Sketch maintains fixed-length shared counters for frequency estimation. It randomly hashes each newly arrived key into some counters. To ensure unbiased estimation, each key is assigned a random coefficient \(\pm 1\) during insertion. To mitigate the variance induced by hash collision, the method employs multiple independent hash functions and uses the median of their outputs as the result of queries. Subsequent work, such as Elastic Sketch \citep{elasticsketch} introduced a frequency-aware key separation mechanism: When a key is detected to exceed predefined size thresholds, it is extracted from the shared counters and stored individually using precise tracking structures. This hybrid approach significantly enhances both storage efficiency and measurement accuracy.

\postsec

\presec
    \section{Contextual Amnesia}
    \label{sec:amnesia}

\begin{figure*}[h]
\centering
    \includegraphics[width=0.7\textwidth]{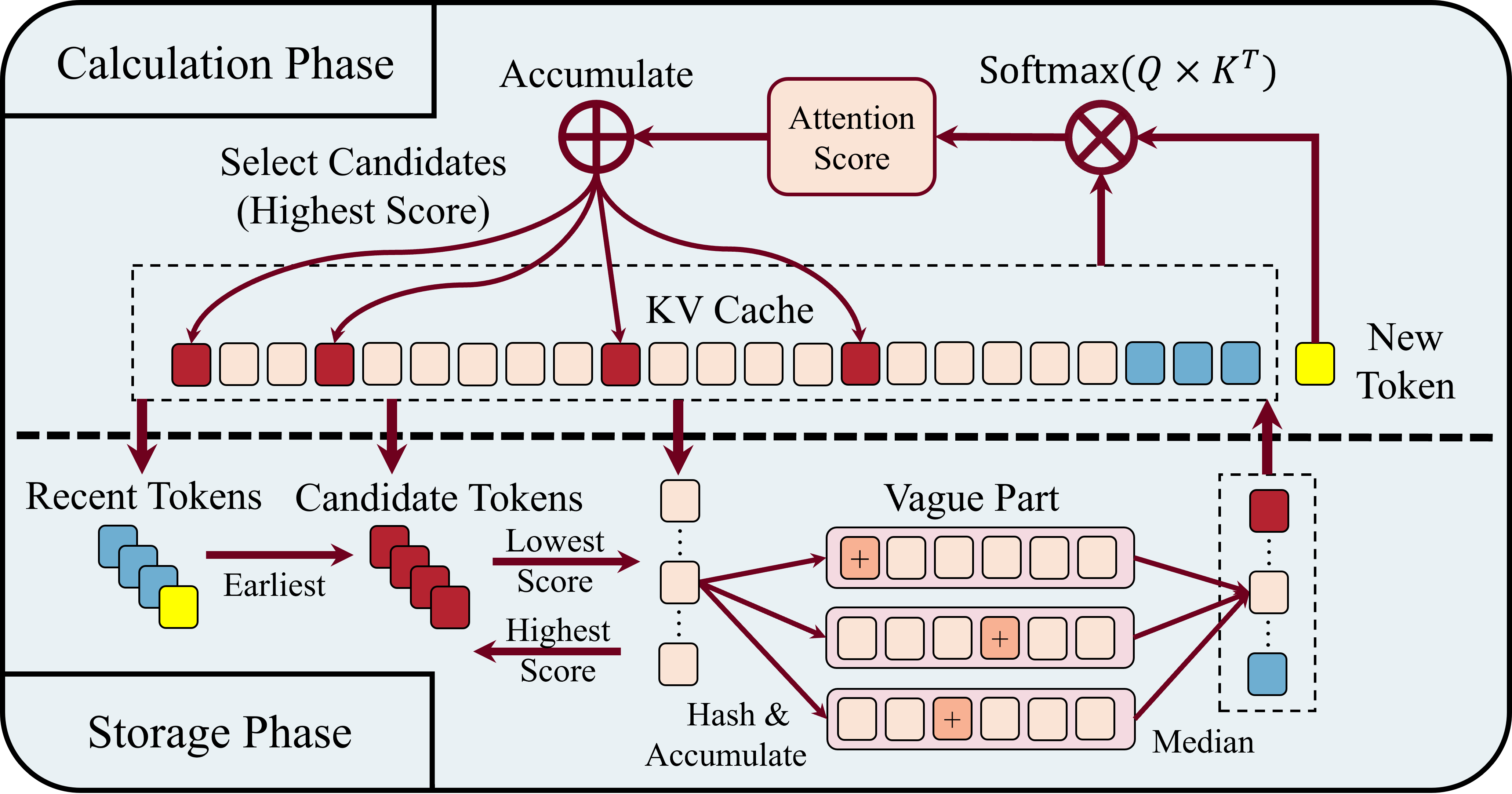}
\caption{Overview of \aname{} system.}
\label{fig:algo:system}
\end{figure*}

    \paragraph{Rethinking Token Sparsity} As is mentioned in Introduction\ref{sec:intro}, token compression methods, based mainly on eviction and merging, almost entirely rely on the assumption that using only a subset of token information can yield a similar result to full attention. However, MagicPig \citep{magicpig} has experimentally demonstrated that this sparsity assumption does not hold for certain tasks, such as information-gathering tasks like word counting. This leads to the emergence of Contextual Amnesia.
    
Does this mean Contextual Amnesia will not occur for the remaining tasks? Unfortunately, the answer is no. Even when the sparsity assumption holds, these methods still require accurate prediction of tokens with high future attention scores. Almost all current prediction methods are history-based, assuming that tokens with high cumulative historical attention scores will also have high scores in the future. However, this approach has inherent problems.

We conducted a 2k-context text summarization task on Llama-2-7B-Chat \citep{llama2}, recording all attention scores, as is shown in Figure \ref{fig:amnesia_exp}. We then simulated the Heavy Hitter (HH) calculation methods of H2O \citep{h2o} and SnapKV \citep{snapkv} (excluding their Recent component), with the simulated HH shown in dark red in the graph. Attention scores of non-HH were calculated and marked in yellow. We defined a token as ``activated'' when its attention score exceeded 0.5\% (a relatively high threshold considering sink tokens often account for over 90\% of attention scores), and counted each token's activation times. Figure \ref{fig:amnesia_exp} shows the attention score distributions and activation counts of H2O and SnapKV under different HH Ratios.
Notably, even with a HH Ratio of 0.4, many non-HH tokens still exhibit high attention scores and frequent activations. This indicates that existing prediction methods are often inaccurate. Even when the sparsity assumption is valid, they can still lead to Contextual Amnesia, proving the necessity of \aname{}.


    
\postsec

\presec
    \section{Algorithm}
    \label{sec:algo}

\subsection{Overview}

\aname{} is deployed independently on each of the attention layers in the Decoder-only models, and is triggered when the layer is involved in computation during the forward pass. Figure \ref{fig:algo:system} illustrates the overall framework of \aname{} within one layer.

When this layer is not computed (Storage Phase), tokens are stored statically in three different parts: Recent, Candidate, and Vague. Recent Part contains the newest generated tokens, and Candidate Part keeps the ``most important'' tokens selected by accumulative attention scores. Both of them store raw tokens that are directly derived from the KV cache without compression. The other tokens are compressed and stored in a modified Count Sketch named Vague Part as the sketch mentioned in Background\ref{sec:background}, which supports token insertion and query in batches. 

When this layer is computed (Calculation Phase), we first rebuild the KV cache from the storage. For Recent and Candidate tokens that are stored directly, we only need to reuse them. As for Vague tokens, we need to make batched queries on the sketch to reconstruct them. After rebuilding the KV cache, the self-attention computation proceeds directly, and we only need to retain and accumulate the attention scores for each token, which is the result of \(\mathrm{Softmax}(QK^T)\). Finally, we only need to insert the newly generated token into the storage.

One can note that, only one layer can be activated at a time. That is to say, at most one out of tens of layers keeps the memory of its full KV cache. Therefore, the process of rebuilding the KV cache has only a minor impact on peak memory usage.



\SetKwProg{Fn}{Function}{}{}
\SetKwFunction{FnCompress}{Compress}
\SetKwFunction{FnRevive}{Revive}
\SetKwFunction{FnInsert}{V.Insert}
\SetKwFunction{FnQuery}{V.Query}

\SetKwFunction{FInsert}{Insert}

\subsection{Vague Sketch}

The Vague Part consists of $r$ buckets, each containing $b$ tensors with the same shape as a token's Embedding as its elements. The tensors are shared for all tokens in the Vague Part, and initially all tensors are zeros. In the example of Figure \ref{fig:algo:system}, we can see that $r=3, b=6$.

Algorithm \ref{alg:algo1} and \ref{alg:algo2} outline the main insertion and query procedures in the Vague Part. To insert tokens, we first employ $r$ distinct hash functions $h_1, \ldots, h_r$ that map each token to $r$ indices, representing its positions in the $r$ buckets. For each mapped position, we accumulate the tokens' key and value embedding as their compressed estimation. To make queries of certain tokens, we use the same hash functions to fetch the corresponding hashed positions' estimation, and use their medians as the query result.

One can see that the Vague Part only uses memory of \(O(rb)\) tokens, regardless of the actual number of tokens inserted. Since hash collisions become unavoidable when a sufficiently large number of tokens are inserted, the compression of the Vague Part inevitably introduces additional errors. According to information entropy theory, error-free efficient compression is impossible. Nevertheless, we have taken various measures to minimize the magnitude of these errors.

One approach, as mentioned earlier, is to take the median during querying to reduce the standard deviation caused by hash collisions. The second approach involves randomly multiplying the value embeddings by coefficients of \(\pm 1\) during both insertion and query operations (\(g_i\) in pseudocodes). This technique aims to make the expected bias to zero, rendering the estimation of values unbiased. It is worth noting that key embeddings do not require such unbiasedness, as the subsequent computation of \(\mathrm{Softmax}(QK^T)\) involves a nonlinear Softmax operation, which cannot preserve unbiasedness after transformation. The additional error bounds of \aname{} are analyzed in the Math section\ref{sec:math}.



\begin{algorithm}
\SetAlgoLined
\caption{Insert token to Vague sketch}
\KwIn{Vague sketch \(V\), token T}
\tcp*{Supports batched insertions}
\KwOut{Updated Vague sketch \(V\)}
\SetKwFunction{FIsInVague}{Is\_In\_Vague}
\SetKwFunction{FIsInRecent}{Is\_In\_Recent}

\BlankLine
\Fn{
\FnInsert{$V$, $T$}
}{
    
    $J \leftarrow T.\text{index}$\;
    
    \For{$i = 1$ \KwTo $r$}{
        $V[i][h_i(J)].\text{key} \leftarrow V[i][[h_i(J)].\text{key} + T.\text{key}$\;
        \tcp*{Keys and values are vectors}
        $V[i][[h_i(J)].\text{value} \leftarrow V[i][[h_i(J)].\text{value} + T.\text{value}\times g_i(J) $\;
    }
    \Return{$V$}\;
    }
\label{alg:algo1}
\end{algorithm}

\begin{algorithm}
\SetAlgoLined
\caption{Query token from Vague sketch}
\KwIn{Vague sketch $V$, Token Index $J$} \tcp*{Supports batched queries}
\KwOut{Token Embedding $\langle$ key, value$\rangle$}

\BlankLine
\Fn{\FnQuery{$V$, $J$}}{ 
    $res \leftarrow []$\;
    
    \For{$i = 1$ \KwTo $r$}{
        $res$.append($\langle V[i][h_i(J)].\text{key}, V[i][h_i(J)].\text{value} \times g_i(J) \rangle$)\;
    }
    \Return{\text{Median}($res$, dim=0)}\;
}
\label{alg:algo2}

\end{algorithm}

\subsection{Recent and Candidate}

Inspired by Elastic Sketch \citep{elasticsketch}, we aim to introduce frequency-aware token separation into the Vague Sketch. However, since each token arrives only once when viewed as a data stream, we need a separation strategy beyond frequency. Coincidentally, the cumulative attention scores used in traditional KV cache compression methods align perfectly with the frequency-aware token separation concept. Therefore, we implement similar Recent and Candidate components to specifically store the newest tokens and those with the highest attention scores. Unlike traditional KV cache compression methods, however, \aname{} inserts tokens that would otherwise be evicted or merged into the Vague Part for potential reconstruction when needed.

\begin{algorithm}
\SetAlgoLined
\caption{Insert newly generated token(s) \(T\) and compress}
\KwIn{Current Storage $\langle R, C, V \rangle$, new token(s) $T$} \tcp* {\(R\) works as queue, \(C\) works as min-heap}
\KwOut{Updated Storage $\langle R, C, V \rangle$}

\BlankLine
\Fn{
\FnCompress{$\langle R, C, V \rangle$, $T$}
}{
    $R$.Push($T$)\;
    
    \If{$R$.size $> R$.budget + Slack}{
        
        $C$.Push($R$.Pop\_k($R$.size $- R$.budget))        
    }
    \While{$C$.size $> C$.budget}{
        $V$.Insert($C$.Top\_Pop())
    }
    \While{$C$.Min\_Score() $\times$ ReplaceRate $< V$.Max\_Score()}{
        tmp $\leftarrow V$.Query($V$.Max\_Score\_Index())\;
        
        $V$.Delete(tmp)\;
        
        $V$.Insert($C$.Top\_Pop())\;
        
        $C$.Push(tmp)\;
    }
    \Return{$\langle R, C, V \rangle$}\;
    }
\label{alg:algo3}

\end{algorithm}

\begin{algorithm}
\SetAlgoLined
\caption{Rebuild the KV Cache}
\KwIn{Storage $\langle R, C, V \rangle$} 
\KwOut{Rebuilt KV Cache \(KV\)}
\SetKwFunction{FIsInVague}{Is\_In\_Vague}
\SetKwFunction{FIsInRecent}{Is\_In\_Recent}

\BlankLine
\Fn{
\FnRevive{$\langle R, C, V \rangle$}
}{
    $KV \leftarrow \text{Zeros}(N)$\;
    
    \For{$i = 1$ \KwTo $N$}{
        \If{\FIsInVague{$i$}}{
            $KV[i] \leftarrow V$.query($i$)\;
        }
        \ElseIf{\FIsInRecent{$i$}}{
            $KV[i] \leftarrow R$.get($i$)\;
        }
        \Else{
            $KV[i] \leftarrow C$.get($i$)\;
        }
    }
    \Return{$KV$}\;
    }
\label{alg:algo4}

\end{algorithm}

Algorithms \ref{alg:algo3} and \ref{alg:algo4} illustrate the complete compression and rebuild processes. When new tokens arrive, we first push them to the Recent Part. The Recent Part operates similarly to a queue. When its size exceeds the allocated budget over a predefined slack value, it pops the oldest tokens from the tail end in bulk and pushes them into the Candidate Part. The Candidate Part operates like a min-heap. It keeps popping tokens with minimum attention scores to the Vague Part. Additionally, we set a swap mechanism, allowing tokens in the Vague Part to be swapped back to the Candidate Part if it's significant enough. For rebuilding, we only need to reuse tokens from Recent and Candidate Part, and query the others from Vague Part.

It is important to note that we have implemented parallelized Vague Part insertion and query operations using torch, so the loops shown here are for illustrative purposes only. All loops are replaced with high-performance tensor operations in our code implementation.    

\postsec

\presec
    \section{Mathematical Analysis}
    \label{sec:math}
\postsec

In this section, we aim to give a theoretical error bound of \aname{}. However, due to the nonlinear nature of the softmax layer, which can exponentially amplify minor perturbations, accurate error estimation is challenging without additional assumptions. Hence, we make the following assumptions.

\begin{itemize}
    \item The bucket number of Vague Part, \(r=3\) holds, which is very common in practice.
    \item Each element of the Query, Key and Value embeddings is independently and identically distributed following a normal distribution, that is, \(Q \sim \mathcal{N}(0, \sigma_q), K \sim \mathcal{N}(0, \sigma_k), V \sim \mathcal{N}(0, \sigma_v) \).
\end{itemize}

Although the second assumption appears strong, it has been experimentally measured in \citep{atom} that the activations during model inference are largely concentrated near zero, with the exception of a few outliers. Therefore, we argue that this assumption is reasonable.

Under these assumptions, we can estimate error bounds for the Key and Value embeddings. Let \(K_{ij}\) and \(V_{ij}\) represent arbitrary reconstructed elements of the Key and Value embeddings, respectively. It follows that:

\begin{align*}
P\left(\text{Var}(\text{median of } K_{ij}) > \frac{a\pi}{N} \sigma_k^2\right) &< \exp\left(-\frac{a}{N}\right) \\
P\left(\text{Var}(\text{median of } V_{ij}) > \frac{a\pi}{N} \sigma_v^2\right) &< \exp\left(-\frac{a}{N}\right)
\end{align*}

Here \(a\) denotes number of tokens compressed in Vague Part, and \(N=rb\) denotes the size of Vague Part.

We can consequently define the additive noise as $\Delta K \sim N(0, \sigma_{k_new}^2)$ and $\Delta V \sim N(0, \sigma_{v_new}^2)$, whose variances are estimated above.

Let \(p = \mathrm{Softmax}\left(\frac{Q K^T}{\sqrt{d_k}}\right)\) be the attention score, where $d_k = 128$. Using this formulation, the error in \(p\) is estimated as:

\begin{align*}
\text{Var}(\Delta p_i) &\approx \sigma_A^2 \cdot\left[ p_i^2 (1 - p_i)^2 + \sum_{j \neq i} p_j^2 p_i^2 \right]\\
&=\sigma_q^2 \cdot \sigma_{k\_new}^2 \cdot \left[ p_i^2 (1 - p_i)^2 + \sum_{j \neq i} p_j^2 p_i^2 \right]
\end{align*}

The detailed proof is provided in the Technical Appendix.

\presec
    \section{Experiments}
    \label{sec:exp}

    \begin{figure}[h]
    \centering
    \begin{subfigure}{0.2\textwidth}
        \centering
        \includegraphics[width=\textwidth]{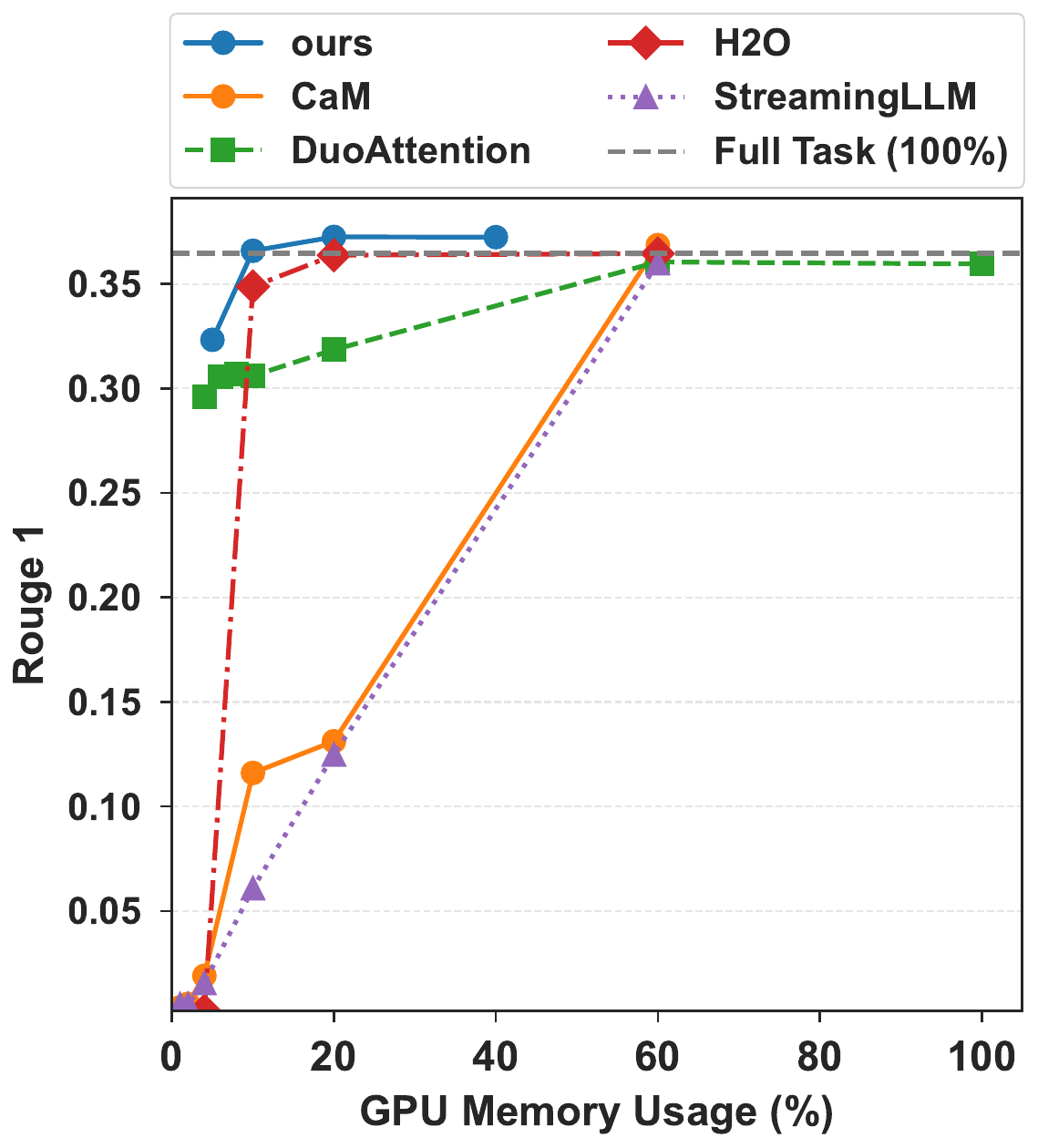}
        \caption{Rouge 1 on XSum.}
        \label{fig:exp:xsum:1}
    \end{subfigure}
    \begin{subfigure}{0.2\textwidth}
        \centering
        \includegraphics[width=\textwidth]{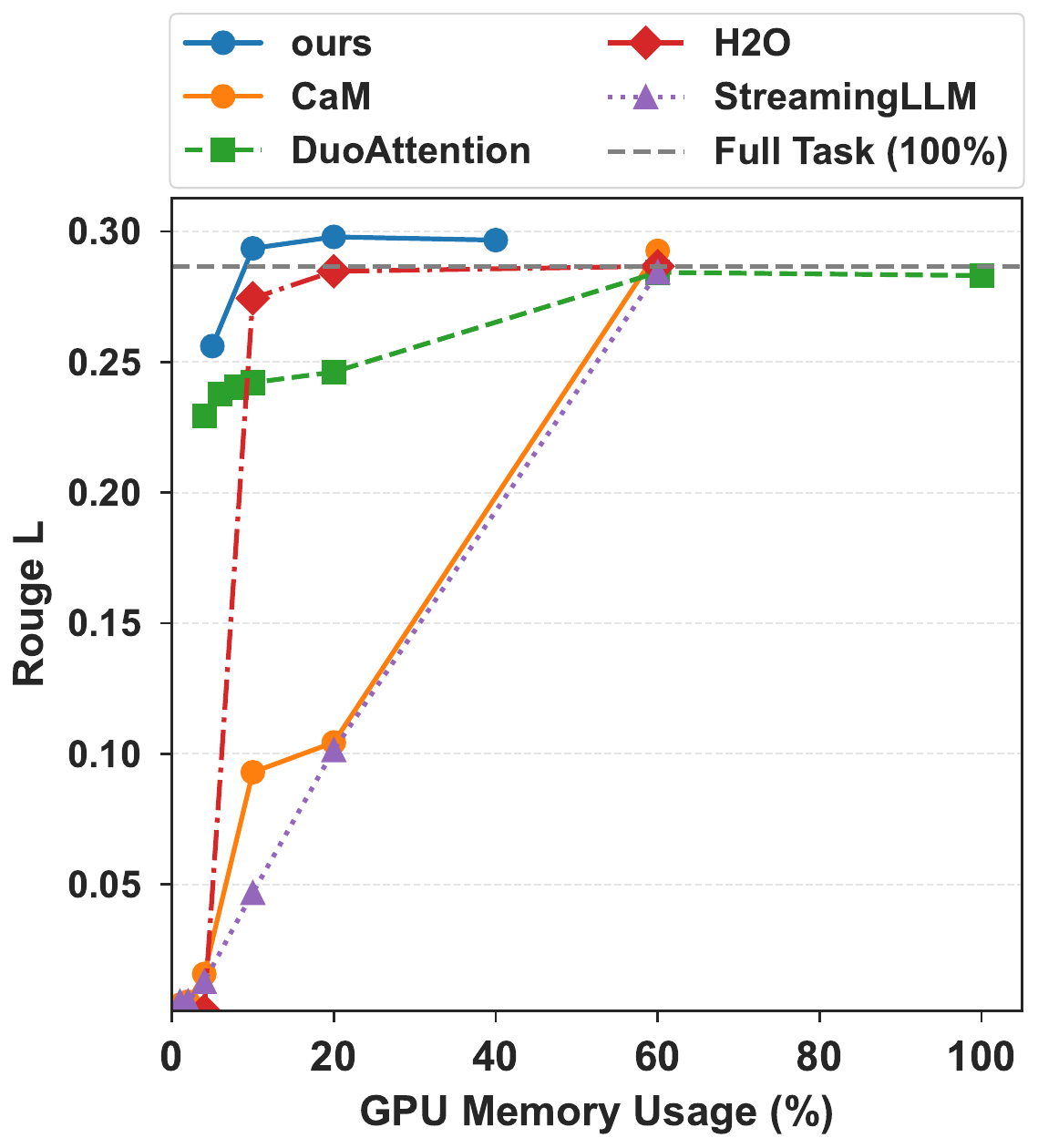}
        \caption{Rouge L on XSum.}
        \label{fig:exp:xsum:2}
    \end{subfigure}
    \hspace{-0.2cm}
    \begin{subfigure}{0.2\textwidth}
        \centering
        \includegraphics[width=\textwidth]{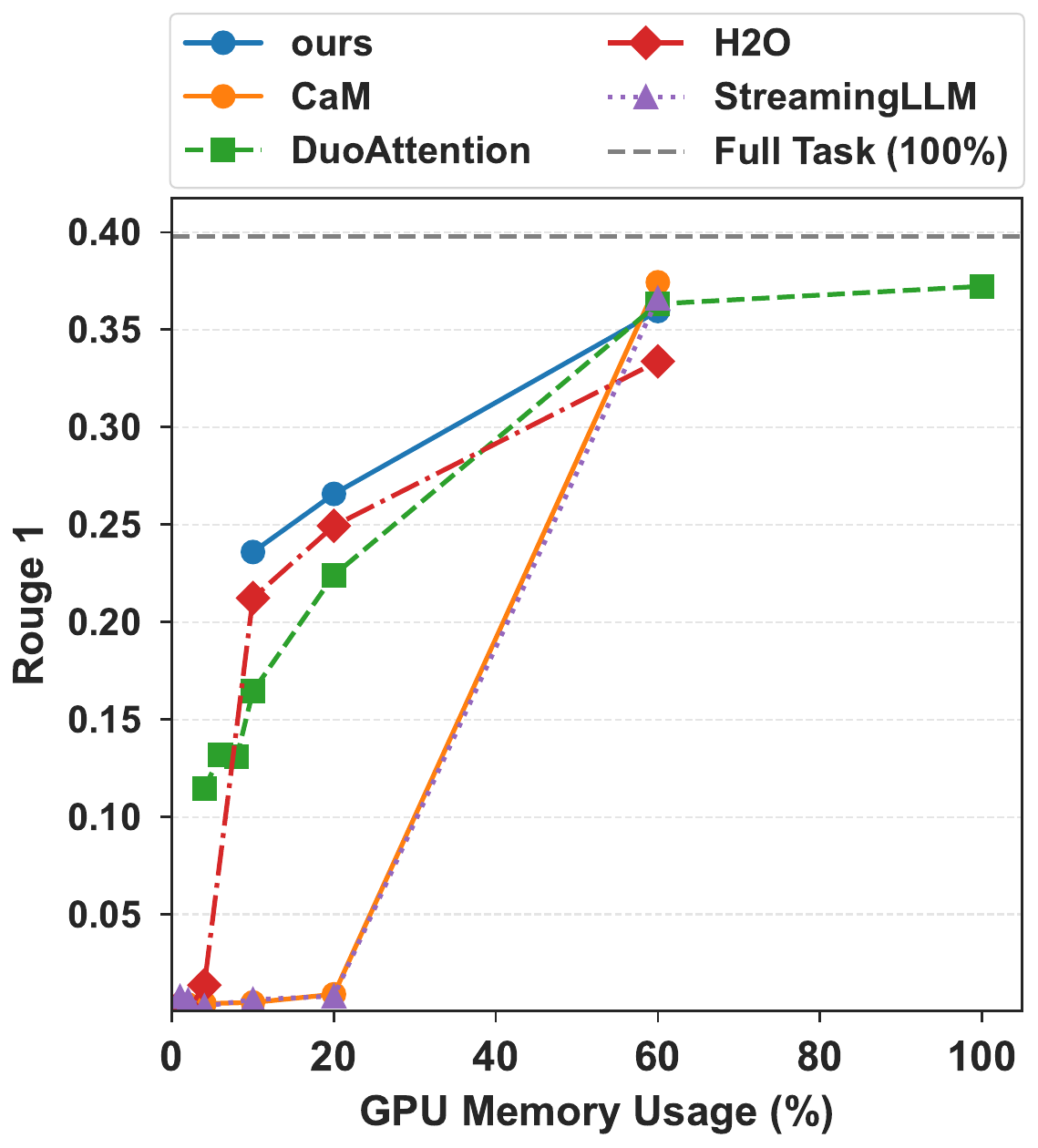}
        \caption{Rouge 1 on CNNDM.}
        \label{fig:exp:cnndm:1}
    \end{subfigure}
    \begin{subfigure}{0.2\textwidth}
        \centering
        \includegraphics[width=\textwidth]{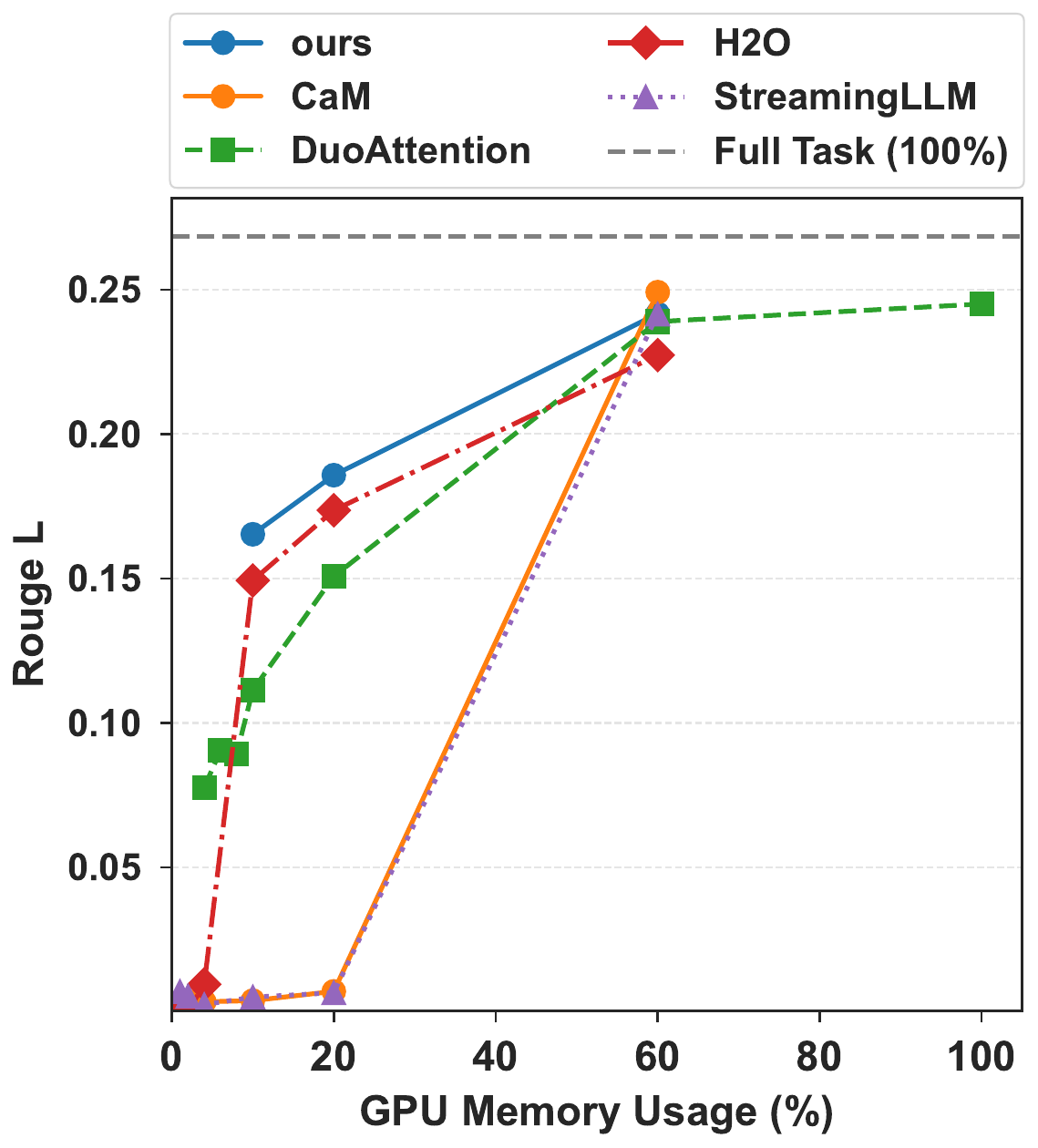}
        \caption{Rouge L on CNNDM.}
        \label{fig:exp:cnndm:2}
    \end{subfigure}
    \caption{Rouge Metrics on short-text datasets.}
    \label{fig:exp:short}
    \end{figure}

    \subsection{Setup}

    \paragraph{Datasets} We employed both short-text and long-text datasets in our experiments to verify that \aname{} can adapt well to various environments.

    Short-text datasets: We selected two datasets from the helm \citep{helm} evaluation framework: XSum \citep{xsum} and CNNDM \citep{cnndm}. These datasets have been modified and synthesized by helm for article summarization tasks, with an average length of approximately 2k tokens.
    
    
    Long-text datasets: Two long-text datasets were chosen: NIAH \citep{niah} and Longbench \citep{longbench}. NIAH involves inserting a ``needle'' (critical information) into an extremely long story and then detecting whether the model can reconstruct this needle’s information. Longbench is a comprehensive dataset encompassing diverse tasks such as document QA, article summarization, few-shot learning, code filling, and more. Both long-text datasets can reach up to 32k tokens in length.

    \paragraph{Models} We selected the popular open-source models Llama-2-7B-chat \citep{llama2} (supporting a 2k context window) and its long-text fine-tuned version Llama-2-7B-32K-Instruct \citep{llama2-7b-32k-instruct} (supporting a 32k context window) as the baseline models for our experiments.

    \begin{figure}[h]
    \centering
    \begin{subfigure}{0.22\textwidth}
        \centering
        \includegraphics[width=\textwidth]{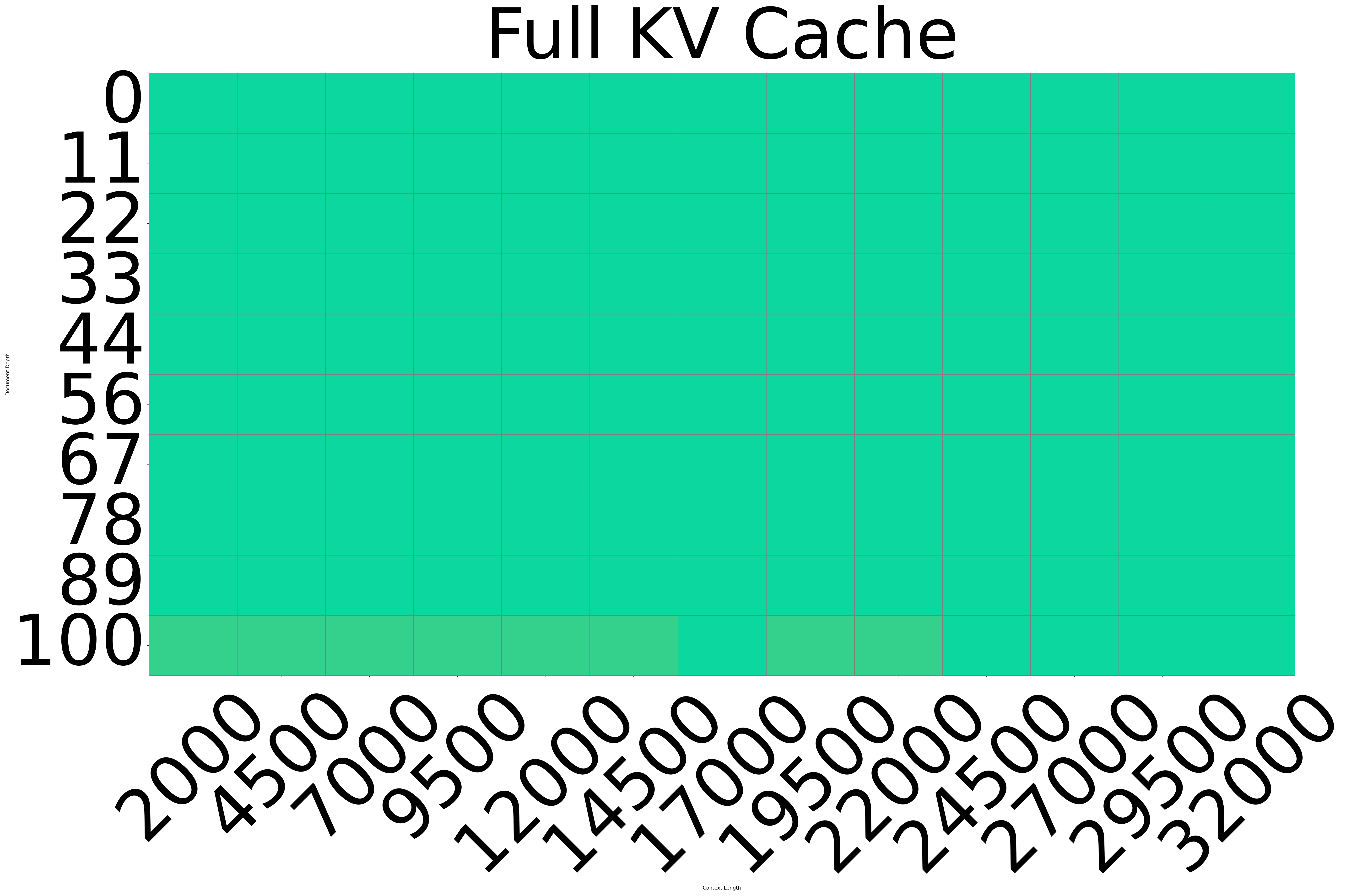}
        \label{fig:app:niah:full}
    \end{subfigure}
    \begin{subfigure}{0.22\textwidth}
        \centering
        \includegraphics[width=\textwidth]{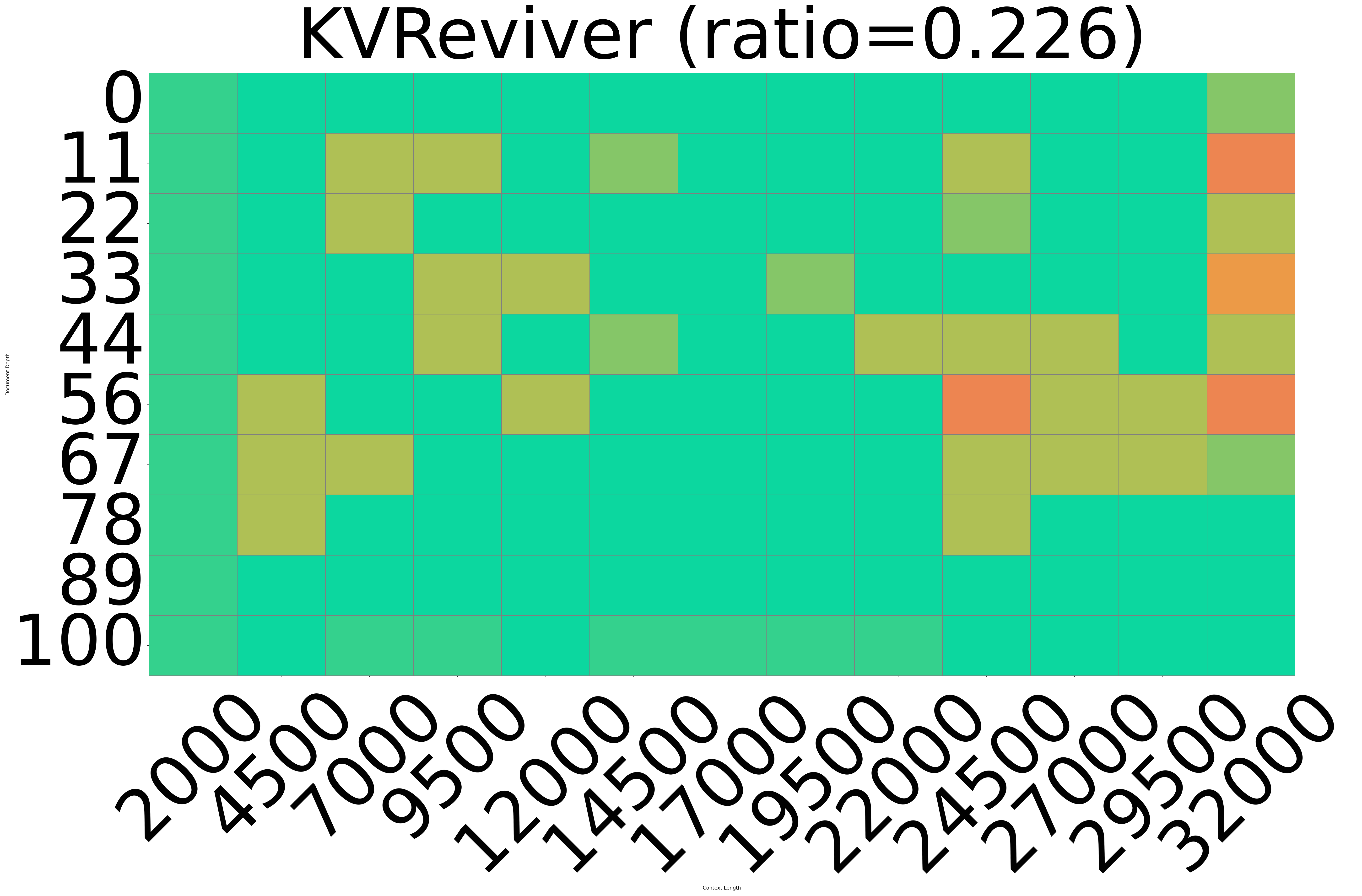}
        \label{fig:app:niah:ours}
    \end{subfigure}
    \begin{subfigure}{0.22\textwidth}
        \centering
        \includegraphics[width=\textwidth]{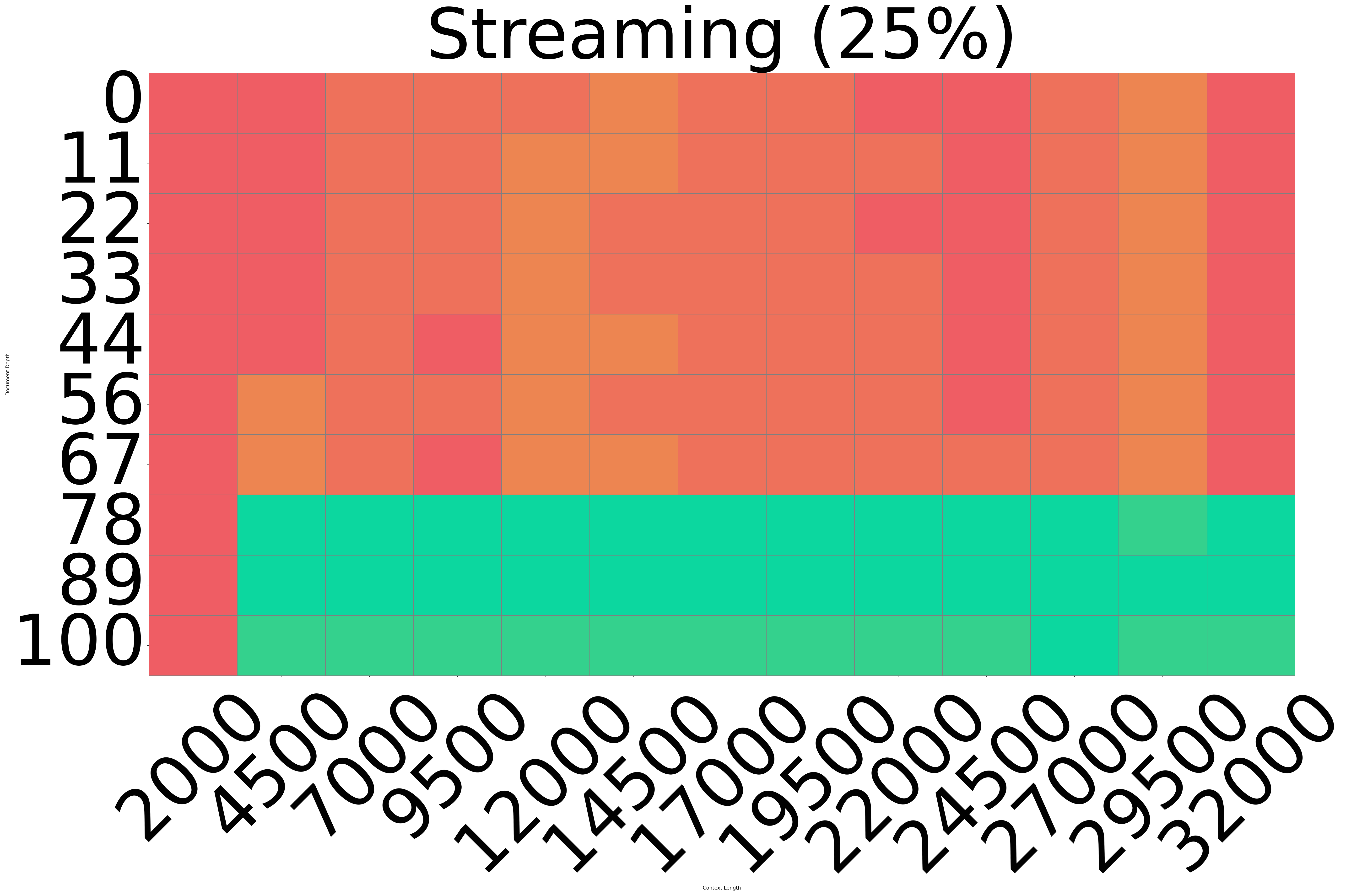}
        \label{fig:app:niah:streaming}
    \end{subfigure}
    \begin{subfigure}{0.22\textwidth}
        \centering
        \includegraphics[width=\textwidth]{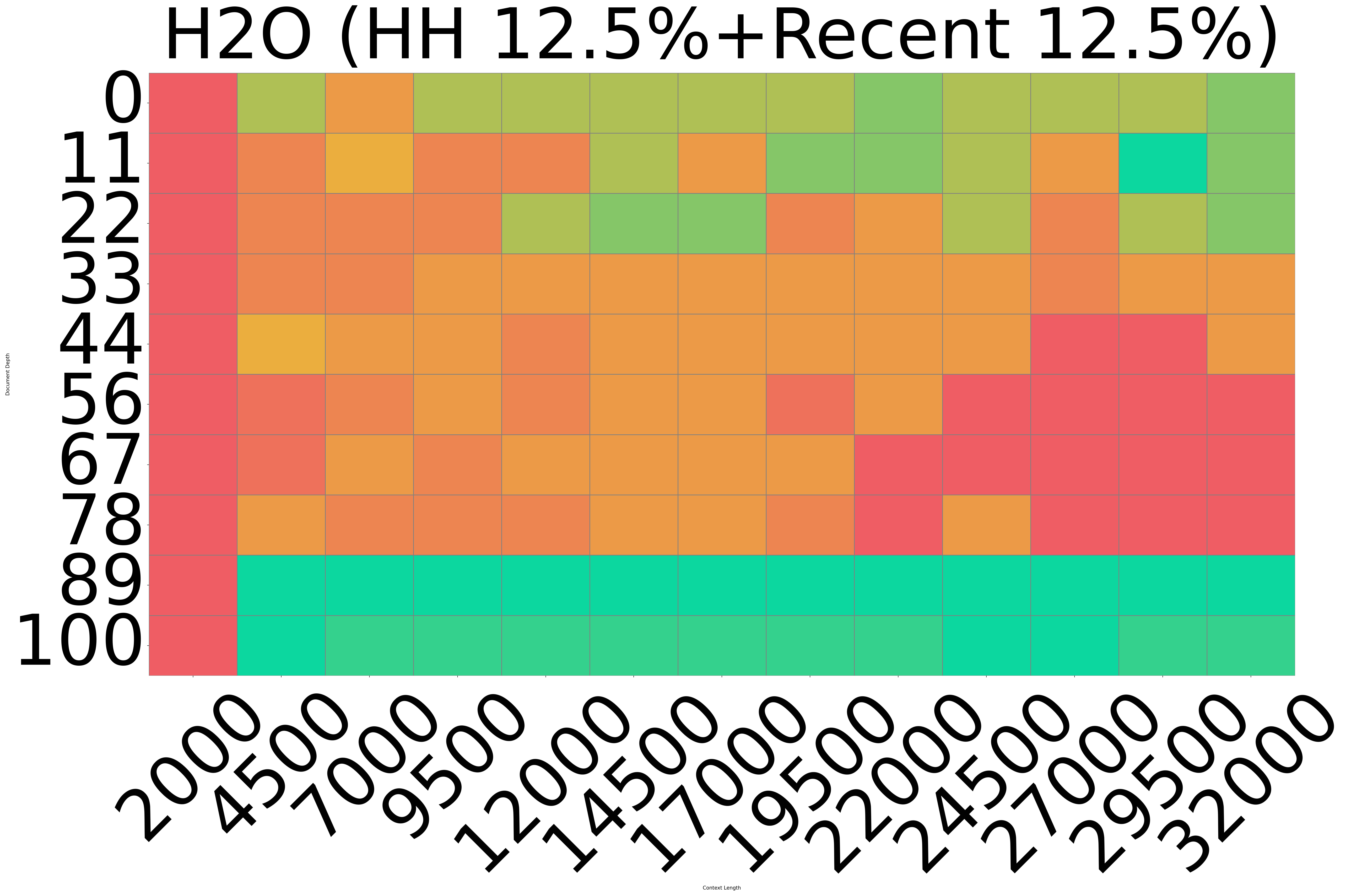}
        \label{fig:app:niah:h2o}
    \end{subfigure}
    \begin{subfigure}{0.22\textwidth}
        \centering
        \includegraphics[width=\textwidth]{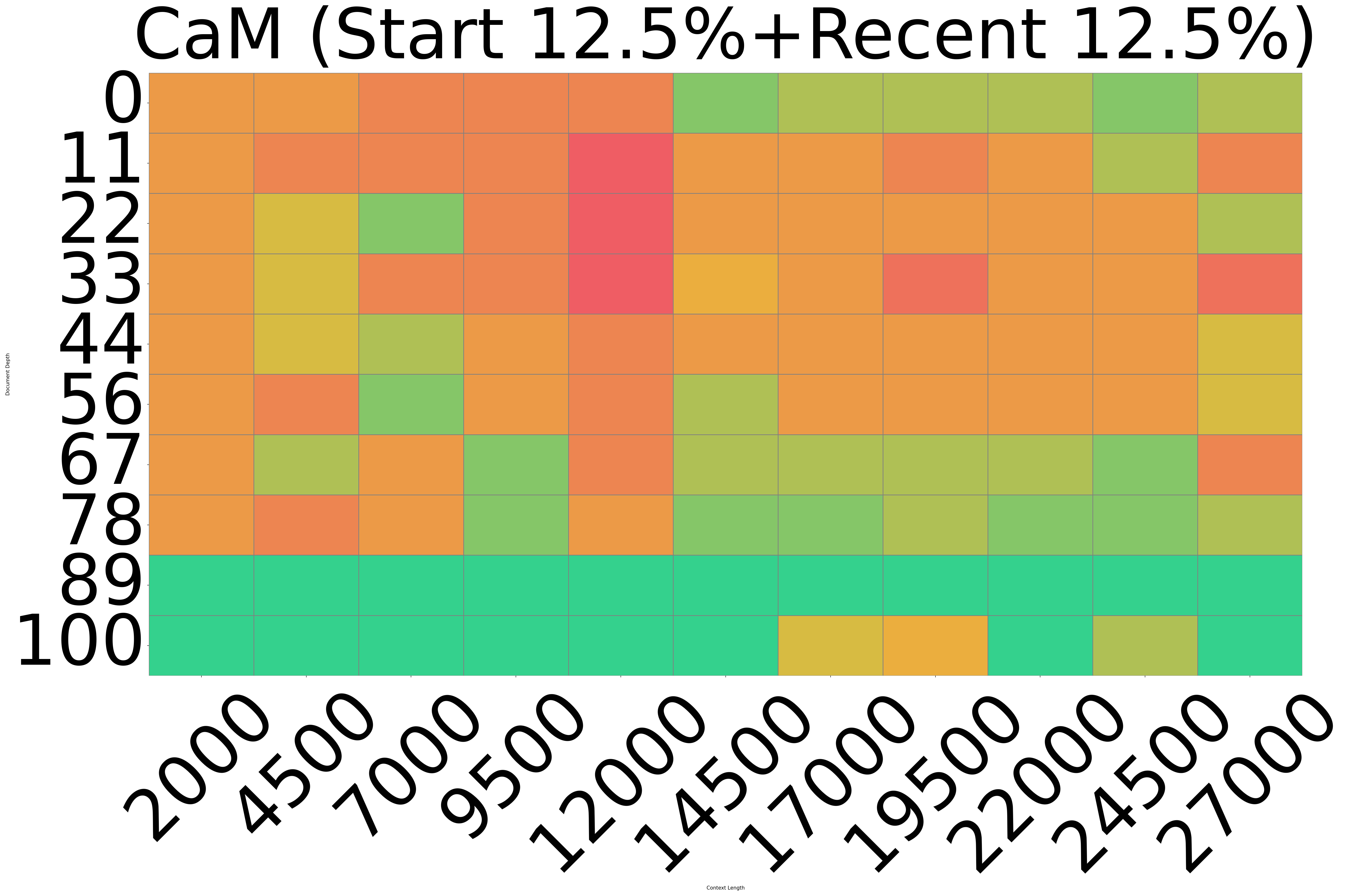}
        \label{fig:app:niah:cam}
    \end{subfigure}
    \caption{NIAH Results on different methods with around 25\% KV Cache budgets.}
    \label{fig:app:naih}
    \end{figure}

    \begin{figure}[htbp]
    \centering
    \includegraphics[width=1\linewidth]{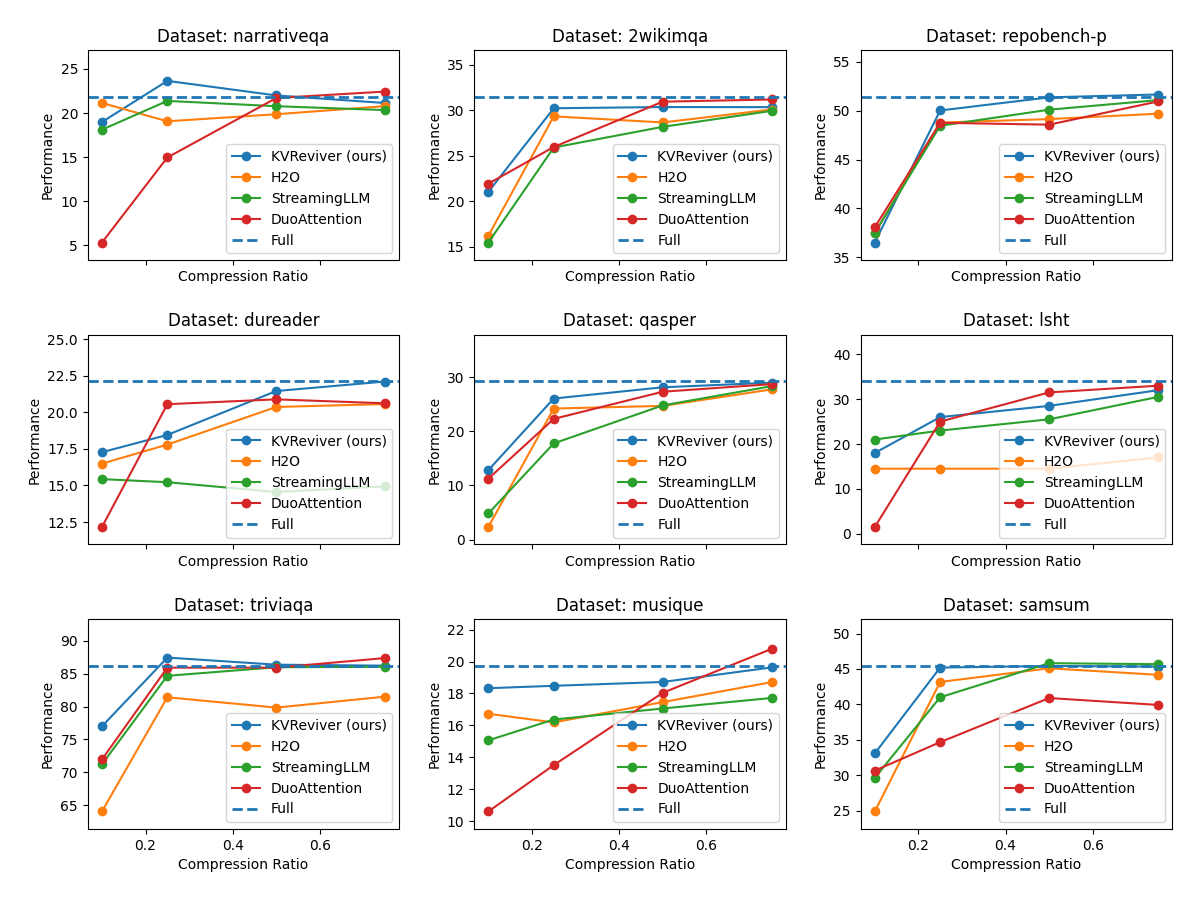}
    \caption{\aname{} showed better space-accuracy trade-off than related works.}
    \label{fig:exp:longbench}
    \end{figure}

    \paragraph{Comparison Baselines} We selected well-established KV Cache eviction methods, including H2O \citep{h2o} and StreamingLLM \citep{streamingllm}, and the merging method CaM \citep{cam} as comparison baselines. For certain tasks, we also compare with DuoAttention \citep{duoattention}, although DuoAttention is fundamentally compatible with our approach.

    \paragraph{Implementations} All experiments in this paper were conducted on a server with 8 NVIDIA A100-SXM4-80GB GPUs. The core code is implemented using torch, with all core operations, including compression-revival, adopting tensorized implementations to minimize the time overhead introduced by additional operations as much as possible.

    \subsection{Result on Short-Text Datasets}

    On the two short-text datasets, our approach consistently outperformed related work, as is shown in Figure \ref{fig:exp:short}. Notably, in the relatively simple XSum task, when the compression ratio was $\geq 10\%$ , the inference accuracy of our method even exceeded that of the full KV Cache. This phenomenon may be attributed to the fact that the full KV Cache itself contains redundant information, while our approach filters out such redundancy, enabling the model to more easily focus on valuable tokens during inference.
    
    \subsection{Result on Long-Text Datasets}
    
    \paragraph{Result on NIAH dataset} As illustrated in Figure \ref{fig:app:naih}, when retaining only approximately 25\% of the memory budget, \aname{} achieves retrieval results comparable to the full-volume baseline across most positions. In contrast, state-of-the-art methods barely manage to retrieve the Needle under the same constraint. This performance advantage is attributed to our compression-restoration mechanism, which enables the successful restoration of tokens corresponding to the Needle, thereby facilitating successful retrieval.
    
    \paragraph{Result on Longbench dataset} As shown in Figure \ref{fig:exp:longbench}, on most subtasks, the proposed method can achieve results comparable to or close to the full-volume version (with approximately 2\% accuracy loss) using only 25\% of the memory budget, outperforming related work. This result is consistent with the findings on the NIAH dataset, fully demonstrating the effectiveness of the proposed method in the compression-revival task and its suitability for handling long-text tasks.



\postsec

\presec
    \section{Conclusion}
    \label{sec:conc}
\postsec


We propose \aname{}, the first reversible KV cache compression mechanism enabling token reconstruction. By integrating sketch data structures with KV cache compression, our approach effectively addresses the prevalent Contextual Amnesia problem inherent in traditional non-reversible compression algorithms. Experimental results demonstrate consistent performance improvements over baseline methods: for 2k-length contexts, \aname{} requires only 10\% of the KV cache budget while maintaining identical end-to-end inference accuracy. For 32k-length contexts, it achieves comparable accuracy(\(\sim\)2\% degradation) using merely 25\% of the KV cache budget.


\clearpage
\bibliography{Body/reference}

@article{streamingllm,
  title={Efficient streaming language models with attention sinks},
  author={Xiao, Guangxuan and Tian, Yuandong and Chen, Beidi and Han, Song and Lewis, Mike},
  journal={arXiv preprint arXiv:2309.17453},
  year={2023}
}

@article{h2o,
  title={H2o: Heavy-hitter oracle for efficient generative inference of large language models},
  author={Zhang, Zhenyu and Sheng, Ying and Zhou, Tianyi and Chen, Tianlong and Zheng, Lianmin and Cai, Ruisi and Song, Zhao and Tian, Yuandong and R{\'e}, Christopher and Barrett, Clark and others},
  journal={Advances in Neural Information Processing Systems},
  volume={36},
  pages={34661--34710},
  year={2023}
}

@article{snapkv,
  title={Snapkv: Llm knows what you are looking for before generation},
  author={Li, Yuhong and Huang, Yingbing and Yang, Bowen and Venkitesh, Bharat and Locatelli, Acyr and Ye, Hanchen and Cai, Tianle and Lewis, Patrick and Chen, Deming},
  journal={Advances in Neural Information Processing Systems},
  volume={37},
  pages={22947--22970},
  year={2024}
}

@article{pyramidinfer,
  title={Pyramidinfer: Pyramid kv cache compression for high-throughput llm inference},
  author={Yang, Dongjie and Han, XiaoDong and Gao, Yan and Hu, Yao and Zhang, Shilin and Zhao, Hai},
  journal={arXiv preprint arXiv:2405.12532},
  year={2024}
}

@article{pyramidkv,
  title={Pyramidkv: Dynamic kv cache compression based on pyramidal information funneling},
  author={Cai, Zefan and Zhang, Yichi and Gao, Bofei and Liu, Yuliang and Liu, Tianyu and Lu, Keming and Xiong, Wayne and Dong, Yue and Chang, Baobao and Hu, Junjie and others},
  journal={arXiv preprint arXiv:2406.02069},
  year={2024}
}

@inproceedings{cam,
  title={Cam: Cache merging for memory-efficient llms inference},
  author={Zhang, Yuxin and Du, Yuxuan and Luo, Gen and Zhong, Yunshan and Zhang, Zhenyu and Liu, Shiwei and Ji, Rongrong},
  booktitle={Forty-first International Conference on Machine Learning},
  year={2024}
}

@article{dmc,
  title={Dynamic memory compression: Retrofitting llms for accelerated inference},
  author={Nawrot, Piotr and {\L}a{\'n}cucki, Adrian and Chochowski, Marcin and Tarjan, David and Ponti, Edoardo M},
  journal={arXiv preprint arXiv:2403.09636},
  year={2024}
}

@article{d2o,
  title={D2o: Dynamic discriminative operations for efficient generative inference of large language models},
  author={Wan, Zhongwei and Wu, Xinjian and Zhang, Yu and Xin, Yi and Tao, Chaofan and Zhu, Zhihong and Wang, Xin and Luo, Siqi and Xiong, Jing and Zhang, Mi},
  journal={arXiv preprint arXiv:2406.13035},
  year={2024}
}

@inproceedings{countsketch,
  title={Finding frequent items in data streams},
  author={Charikar, Moses and Chen, Kevin and Farach-Colton, Martin},
  booktitle={International colloquium on automata, languages, and programming},
  pages={693--703},
  year={2002},
  organization={Springer}
}

@article{magicpig,
  title={Magicpig: Lsh sampling for efficient llm generation},
  author={Chen, Zhuoming and Sadhukhan, Ranajoy and Ye, Zihao and Zhou, Yang and Zhang, Jianyu and Nolte, Niklas and Tian, Yuandong and Douze, Matthijs and Bottou, Leon and Jia, Zhihao and others},
  journal={arXiv preprint arXiv:2410.16179},
  year={2024}
}

@article{fisher,
  title={A tutorial on Fisher information},
  author={Ly, Alexander and Marsman, Maarten and Verhagen, Josine and Grasman, Raoul PPP and Wagenmakers, Eric-Jan},
  journal={Journal of Mathematical Psychology},
  volume={80},
  pages={40--55},
  year={2017},
  publisher={Elsevier}
}

@article{deepseek-v2,
  title={Deepseek-v2: A strong, economical, and efficient mixture-of-experts language model},
  author={Liu, Aixin and Feng, Bei and Wang, Bin and Wang, Bingxuan and Liu, Bo and Zhao, Chenggang and Dengr, Chengqi and Ruan, Chong and Dai, Damai and Guo, Daya and others},
  journal={arXiv preprint arXiv:2405.04434},
  year={2024}
}

@inproceedings{cla,
  title={Reducing transformer key-value cache size with cross-layer attention},
  author={Brandon, William and Mishra, Mayank and Nrusimha, Aniruddha and Panda, Rameswar and Ragan-Kelley, Jonathan},
  booktitle={The Thirty-eighth Annual Conference on Neural Information Processing Systems},
  year={2024}
}

@article{mqa,
  title={Fast transformer decoding: One write-head is all you need},
  author={Shazeer, Noam},
  journal={arXiv preprint arXiv:1911.02150},
  year={2019}
}

@article{gqa,
  title={Gqa: Training generalized multi-query transformer models from multi-head checkpoints},
  author={Ainslie, Joshua and Lee-Thorp, James and De Jong, Michiel and Zemlyanskiy, Yury and Lebr{\'o}n, Federico and Sanghai, Sumit},
  journal={arXiv preprint arXiv:2305.13245},
  year={2023}
}

@article{seed-thinking-v1.5,
  title={Seed-thinking-v1. 5: Advancing superb reasoning models with reinforcement learning},
  author={Seed, ByteDance and Yuan, Yufeng and Yue, Yu and Wang, Mingxuan and Zuo, Xiaochen and Chen, Jiaze and Yan, Lin and Xu, Wenyuan and Zhang, Chi and Liu, Xin and others},
  journal={arXiv preprint arXiv:2504.13914},
  year={2025}
}

@inproceedings{palu,
  title={Palu: KV-Cache Compression with Low-Rank Projection},
  author={Chang, Chi-Chih and Lin, Wei-Cheng and Lin, Chien-Yu and Chen, Chong-Yan and Hu, Yu-Fang and Wang, Pei-Shuo and Huang, Ning-Chi and Ceze, Luis and Abdelfattah, Mohamed S and Wu, Kai-Chiang},
  booktitle={The Thirteenth International Conference on Learning Representations},
  year={2024}
}

@article{transformer,
  title={Attention is all you need},
  author={Vaswani, Ashish and Shazeer, Noam and Parmar, Niki and Uszkoreit, Jakob and Jones, Llion and Gomez, Aidan N and Kaiser, {\L}ukasz and Polosukhin, Illia},
  journal={Advances in neural information processing systems},
  volume={30},
  year={2017}
}

@article{duoattention,
  title={Duoattention: Efficient long-context llm inference with retrieval and streaming heads},
  author={Xiao, Guangxuan and Tang, Jiaming and Zuo, Jingwei and Guo, Junxian and Yang, Shang and Tang, Haotian and Fu, Yao and Han, Song},
  journal={arXiv preprint arXiv:2410.10819},
  year={2024}
}

@article{kvquant,
  title={Kvquant: Towards 10 million context length llm inference with kv cache quantization},
  author={Hooper, Coleman and Kim, Sehoon and Mohammadzadeh, Hiva and Mahoney, Michael W and Shao, Sophia and Keutzer, Kurt and Gholami, Amir},
  journal={Advances in Neural Information Processing Systems},
  volume={37},
  pages={1270--1303},
  year={2024}
}

@article{atom,
  title={Atom: Low-bit quantization for efficient and accurate llm serving, 2024},
  author={Zhao, Yilong and Lin, Chien-Yu and Zhu, Kan and Ye, Zihao and Chen, Lequn and Zheng, Size and Ceze, Luis and Krishnamurthy, Arvind and Chen, Tianqi and Kasikci, Baris},
  journal={URL https://arxiv. org/abs/2310.19102},
  year={2023}
}

@article{llama2,
  title={Llama 2: Open foundation and fine-tuned chat models},
  author={Touvron, Hugo and Martin, Louis and Stone, Kevin and Albert, Peter and Almahairi, Amjad and Babaei, Yasmine and Bashlykov, Nikolay and Batra, Soumya and Bhargava, Prajjwal and Bhosale, Shruti and others},
  journal={arXiv preprint arXiv:2307.09288},
  year={2023}
}

@article{helm,
  title={Holistic evaluation of language models},
  author={Liang, Percy and Bommasani, Rishi and Lee, Tony and Tsipras, Dimitris and Soylu, Dilara and Yasunaga, Michihiro and Zhang, Yian and Narayanan, Deepak and Wu, Yuhuai and Kumar, Ananya and others},
  journal={arXiv preprint arXiv:2211.09110},
  year={2022}
}

@article{xsum,
  title={Don’t give me the details, just the summary},
  author={Narayan, Shashi and Cohen, Shay B and Lapata, Mirella},
  journal={Topic-Aware Convolutional Neural Networks for Extreme Summarization. ArXiv abs/1808.08745},
  year={2018}
}

@article{cnndm,
  title={Abstractive text summarization using sequence-to-sequence rnns and beyond},
  author={Nallapati, Ramesh and Zhou, Bowen and Gulcehre, Caglar and Xiang, Bing and others},
  journal={arXiv preprint arXiv:1602.06023},
  year={2016}
}

@misc{niah,
    title = {LLMTest\_NeedleInAHaystack},
    author = {gkamradt},
    howpublished = {\url{https://github.com/gkamradt/LLMTest_NeedleInAHaystack}},
    year = {2023}, 
}

@article{longbench,
  title={Longbench: A bilingual, multitask benchmark for long context understanding},
  author={Bai, Yushi and Lv, Xin and Zhang, Jiajie and Lyu, Hongchang and Tang, Jiankai and Huang, Zhidian and Du, Zhengxiao and Liu, Xiao and Zeng, Aohan and Hou, Lei and others},
  journal={arXiv preprint arXiv:2308.14508},
  year={2023}
}

@inproceedings{elasticsketch,
  title={Elastic sketch: Adaptive and fast network-wide measurements},
  author={Yang, Tong and Jiang, Jie and Liu, Peng and Huang, Qun and Gong, Junzhi and Zhou, Yang and Miao, Rui and Li, Xiaoming and Uhlig, Steve},
  booktitle={Proceedings of the 2018 Conference of the ACM Special Interest Group on Data Communication},
  pages={561--575},
  year={2018}
}

@misc{llama2-7b-32k-instruct,
  title = {Llama-2-7B-32K-Instruct Model},
  author = {togethercomputer},
  howpublished = {\url{https://huggingface.co/togethercomputer/Llama-2-7B-32K-Instruct}},
  year = {2023},
}

@inproceedings{group-fisher,
  title={Group fisher pruning for practical network compression},
  author={Liu, Liyang and Zhang, Shilong and Kuang, Zhanghui and Zhou, Aojun and Xue, Jing-Hao and Wang, Xinjiang and Chen, Yimin and Yang, Wenming and Liao, Qingmin and Zhang, Wayne},
  booktitle={International Conference on Machine Learning},
  pages={7021--7032},
  year={2021},
  organization={PMLR}
}

@misc{deepseekai2025deepseekr1incentivizingreasoningcapability,
      title={DeepSeek-R1: Incentivizing Reasoning Capability in LLMs via Reinforcement Learning}, 
      author={DeepSeek-AI and Daya Guo and Dejian Yang and Haowei Zhang and Junxiao Song and Ruoyu Zhang and Runxin Xu and Qihao Zhu and Shirong Ma and Peiyi Wang and Xiao Bi and Xiaokang Zhang and Xingkai Yu and Yu Wu and Z. F. Wu and Zhibin Gou and Zhihong Shao and Zhuoshu Li and Ziyi Gao and Aixin Liu and Bing Xue and Bingxuan Wang and Bochao Wu and Bei Feng and Chengda Lu and Chenggang Zhao and Chengqi Deng and Chenyu Zhang and Chong Ruan and Damai Dai and Deli Chen and Dongjie Ji and Erhang Li and Fangyun Lin and Fucong Dai and Fuli Luo and Guangbo Hao and Guanting Chen and Guowei Li and H. Zhang and Han Bao and Hanwei Xu and Haocheng Wang and Honghui Ding and Huajian Xin and Huazuo Gao and Hui Qu and Hui Li and Jianzhong Guo and Jiashi Li and Jiawei Wang and Jingchang Chen and Jingyang Yuan and Junjie Qiu and Junlong Li and J. L. Cai and Jiaqi Ni and Jian Liang and Jin Chen and Kai Dong and Kai Hu and Kaige Gao and Kang Guan and Kexin Huang and Kuai Yu and Lean Wang and Lecong Zhang and Liang Zhao and Litong Wang and Liyue Zhang and Lei Xu and Leyi Xia and Mingchuan Zhang and Minghua Zhang and Minghui Tang and Meng Li and Miaojun Wang and Mingming Li and Ning Tian and Panpan Huang and Peng Zhang and Qiancheng Wang and Qinyu Chen and Qiushi Du and Ruiqi Ge and Ruisong Zhang and Ruizhe Pan and Runji Wang and R. J. Chen and R. L. Jin and Ruyi Chen and Shanghao Lu and Shangyan Zhou and Shanhuang Chen and Shengfeng Ye and Shiyu Wang and Shuiping Yu and Shunfeng Zhou and Shuting Pan and S. S. Li and Shuang Zhou and Shaoqing Wu and Shengfeng Ye and Tao Yun and Tian Pei and Tianyu Sun and T. Wang and Wangding Zeng and Wanjia Zhao and Wen Liu and Wenfeng Liang and Wenjun Gao and Wenqin Yu and Wentao Zhang and W. L. Xiao and Wei An and Xiaodong Liu and Xiaohan Wang and Xiaokang Chen and Xiaotao Nie and Xin Cheng and Xin Liu and Xin Xie and Xingchao Liu and Xinyu Yang and Xinyuan Li and Xuecheng Su and Xuheng Lin and X. Q. Li and Xiangyue Jin and Xiaojin Shen and Xiaosha Chen and Xiaowen Sun and Xiaoxiang Wang and Xinnan Song and Xinyi Zhou and Xianzu Wang and Xinxia Shan and Y. K. Li and Y. Q. Wang and Y. X. Wei and Yang Zhang and Yanhong Xu and Yao Li and Yao Zhao and Yaofeng Sun and Yaohui Wang and Yi Yu and Yichao Zhang and Yifan Shi and Yiliang Xiong and Ying He and Yishi Piao and Yisong Wang and Yixuan Tan and Yiyang Ma and Yiyuan Liu and Yongqiang Guo and Yuan Ou and Yuduan Wang and Yue Gong and Yuheng Zou and Yujia He and Yunfan Xiong and Yuxiang Luo and Yuxiang You and Yuxuan Liu and Yuyang Zhou and Y. X. Zhu and Yanhong Xu and Yanping Huang and Yaohui Li and Yi Zheng and Yuchen Zhu and Yunxian Ma and Ying Tang and Yukun Zha and Yuting Yan and Z. Z. Ren and Zehui Ren and Zhangli Sha and Zhe Fu and Zhean Xu and Zhenda Xie and Zhengyan Zhang and Zhewen Hao and Zhicheng Ma and Zhigang Yan and Zhiyu Wu and Zihui Gu and Zijia Zhu and Zijun Liu and Zilin Li and Ziwei Xie and Ziyang Song and Zizheng Pan and Zhen Huang and Zhipeng Xu and Zhongyu Zhang and Zhen Zhang},
      year={2025},
      eprint={2501.12948},
      archivePrefix={arXiv},
      primaryClass={cs.CL},
      url={https://arxiv.org/abs/2501.12948}, 
}

@article{cmsketch,
title = {An improved data stream summary: the count-min sketch and its applications},
journal = {Journal of Algorithms},
volume = {55},
number = {1},
pages = {58-75},
year = {2005},
issn = {0196-6774},
doi = {https://doi.org/10.1016/j.jalgor.2003.12.001},
url = {https://www.sciencedirect.com/science/article/pii/S0196677403001913},
author = {Graham Cormode and S. Muthukrishnan}
}

@inproceedings{spacesaving,
	title={Efficient computation of frequent and top-k elements in data streams},
	author={Metwally, Ahmed and Agrawal, Divyakant and El Abbadi, Amr},
	booktitle={ICDT},
	year={2005},
	organization={}
}

@inproceedings{cocosketch,
author = {Zhang, Yinda and Liu, Zaoxing and Wang, Ruixin and Yang, Tong and Li, Jizhou and Miao, Ruijie and Liu, Peng and Zhang, Ruwen and Jiang, Junchen},
title = {CocoSketch: high-performance sketch-based measurement over arbitrary partial key query},
year = {2021},
isbn = {9781450383837},
publisher = {Association for Computing Machinery},
address = {New York, NY, USA},
url = {https://doi.org/10.1145/3452296.3472892},
doi = {10.1145/3452296.3472892},
booktitle = {Proceedings of the 2021 ACM SIGCOMM 2021 Conference},
pages = {207–222},
numpages = {16},
location = {Virtual Event, USA},
series = {SIGCOMM '21}
}

@misc{kwon2023efficientmemorymanagementlarge,
      title={Efficient Memory Management for Large Language Model Serving with PagedAttention}, 
      author={Woosuk Kwon and Zhuohan Li and Siyuan Zhuang and Ying Sheng and Lianmin Zheng and Cody Hao Yu and Joseph E. Gonzalez and Hao Zhang and Ion Stoica},
      year={2023},
      eprint={2309.06180},
      archivePrefix={arXiv},
      primaryClass={cs.LG},
      url={https://arxiv.org/abs/2309.06180}, 
}

\clearpage	
    \appendix
\section*{TECHNICAL APPENDIX}
\label{sec:appendix}

\section{Throughput}


We conducted throughput experiments on the Llama-2-7B \citep{llama2} model using a single NVIDIA A100-SXM4-80GB GPU. Input data was sourced from the XSum \citep{xsum} dataset, and we measured the average tokens generated per unit time. The corresponding results are presented in Table \ref{table:throughput}, comparing our method (applied with a 10\% compression ratio) against the unoptimized baseline model. Notably, our approach exhibits slower inference speed than the baseline at lower batch sizes. We consider this result acceptable for two reasons:

\begin{itemize}
    \item Computational Superset: Unlike conventional token-level compression methods, our technique first reconstructs all tokens before performing full self-attention computation. Consequently, the computation required by our method strictly encompasses that of the baseline.
    \item Enabling Larger Batches: Although token reconstruction introduces additional computational overhead, compressing the KV cache allows computation with significantly larger batch sizes. At a 10\% compression ratio, our method supports batch sizes at least 8 times larger than the baseline. Therefore, in practical inference scenarios, our method achieves higher throughput than the baseline.
\end{itemize}


\begin{table}[h]
\centering
\caption{Throughput comparison of \aname{} and the baseline model.}
\label{table:throughput}
\begin{tabular}{|c|c|c|}
\hline
Batch Size & Ours (tokens/s) & Baseline Model (tokens/s) \\
\hline
1          & 6.48         & 12.67                                            \\
\hline
2          & 10.16          & 21.48                                           \\
\hline
8          & 19.10          & -                                             \\
\hline
16         & 22.46         & -                                               \\
\hline
\end{tabular}
\end{table}

\section{Parameter Settings}

\begin{table}[h]
\centering
\caption{Rouge 1 result for different Vague Part ratios.}
\label{table:par1}
\begin{tabular}{|c|c|c|c|}
\hline
Vague Ratio & Candidate Ratio & Recent Ratio & Rouge 1 \\
\hline
0.1 & 0.45 & 0.45 & \bf{35.7} \\
\hline
0.15 & 0.425 & 0.425 & 35.2 \\
\hline
0.2 & 0.4 & 0.4 & 34.4 \\
\hline
0.3 & 0.35 & 0.35 & 34.1 \\
\hline
0.4 & 0.3 & 0.3 & 33.3 \\
\hline
\end{tabular}
\end{table}

\begin{table}[h]
\centering
\caption{Rouge 1 result for different Candidate and Recent Part ratios.}
\label{table:par2}
\begin{tabular}{|c|c|c|c|}
\hline
Vague Ratio & Candidate Ratio & Recent Ratio & Rouge 1 \\
\hline
0.1 & 0.7 & 0.2 & 32.9 \\
\hline
0.1 & 0.6 & 0.3 & 34.5 \\
\hline
0.1 & 0.5 & 0.4 & 35.6 \\
\hline
0.1 & 0.45 & 0.45 & \bf{35.7} \\
\hline
0.1 & 0.4 & 0.5 & 35.0 \\
\hline
0.1 & 0.3 & 0.6 & 35.5 \\
\hline
\end{tabular}
\end{table}

For implementation and mathematical tractability, we fix the number of buckets in the Vague Part to a constant, specifically \(r=3\). This setting is common in sketch-based methods. Given that the total size of the Vague Part is determined by \(rb\), the bucket length \(b\) becomes fixed once \(r\) is set.

Then, we evaluated the impact of the replacement rate coefficient \(\mathrm{ReplaceRate}\), which governs the exchange between the Candidate and Vague parts as outlined in Algorithm 3. Empirical results demonstrate minimal impact when \(\mathrm{ReplaceRate}\) is varied between 1.0 and 1.5. Consequently, we fix this value to 1.1.

This leaves only the relative sizes of the Recent, Candidate, and Vague parts as tunable parameters. Since both the Recent and Candidate parts store tokens precisely, we initially assume equal allocation between them. Under a 10\% compression rate on the XSum task, we enumerated the proportion allocated to the Vague Part. Results (Table \ref{table:par1}) indicate that allocating 0.1 (10\%) to the Vague Part yields optimal performance. Subsequently, we examined whether equal allocation between the Recent and Candidate parts was indeed optimal. Results (Table \ref{table:par2}) confirm that equal allocation provides the best performance.

\section{Limitations}

We have developed \aname{}, a reversible KV cache compression-reconstruction method based on sketch. Despite achieving state-of-the-art performance in both compression ratio and accuracy metrics, there remain several avenues for improvement:

\begin{itemize}
    \item The cumulative attention scoring mechanism poses compatibility challenges with certain inference frameworks, like vLLM \citep{kwon2023efficientmemorymanagementlarge}.
    \item Additional baseline models and comparison algorithms could be incorporated to provide a more comprehensive performance benchmark.
    \item Further optimize the throughput bottleneck to reduce the additional time overhead by an order of magnitude, ensuring that this overhead can be effectively masked by the computational process.
\end{itemize}

\section{Fisher Information}

Mathematically, Fisher Information \citep{fisher} is a statistical measure used to quantify how sensitive a probability distribution is to changes in its parameters, for example, the mean \(\mu\) and standard deviation \(\sigma\) of a normal distribution. Essentially, it characterizes the amount of information carried by each parameter. In machine learning, Fisher Information has been applied to pruning by identifying unimportant parameters \citep{group-fisher}, and Palu \citep{palu} leveraged it to determine the compression rate for rank reduction in each KV head.  


\aname{} computes the Fisher Information of a layer by summing the Fisher Information of the Key and Value weight matrices within the same layer. This value is then used to determine the memory budget allocation ratio across different layers.  


\section{Mathematical Proofs}

Before proving the error bounds mentioned in the main text, we first present a series of lemmas to intuitively illustrate the theoretical justification for the various error-reduction techniques employed.

\begin{lemma}[Variance Reduction by Median]Let $X_1, X_2, X_3$ be i.i.d.\ random variables with distribution $F$ and finite variance $\sigma^2$. Define the median estimator as
\[
\hat{X} := \operatorname{median}(X_1, X_2, X_3).
\]
For any symmetric distribution $F$, we have the following result.
\[
\operatorname{Var}(\hat{X}) \leq \operatorname{Var}(X_1),
\]
with equality if and only if $X_i$ are almost surely constant.
\label{math:lemma1}
\end{lemma}

\begin{proof}
Let $f$ be the density function and $F$ the cumulative distribution function (cdf) of $X_i$. The density of the median is
\[
f_{\hat{X}}(x) = 3F(x)(1-F(x))f(x).
\]
The variance can be written as
\[
\operatorname{Var}(\hat{X}) = \mathbb{E}\big[(\hat{X}-\mu)^2\big] - \big(\mathbb{E}[\hat{X}-\mu]\big)^2,
\]
where $\mu$ is the mean. For symmetric distributions, the bias term vanishes,
\[
\mathbb{E}[\hat{X}-\mu] = 0,
\]
so
\[
\operatorname{Var}(\hat{X}) = \int_{-\infty}^{\infty} (x-\mu)^2 \, 3F(x)(1-F(x))f(x) \, dx.
\]
Comparing with the variance of each $X_i$,
\[
\operatorname{Var}(X_1) = \int_{-\infty}^{\infty} (x-\mu)^2 f(x) \, dx,
\]
and since $0 \leq 3F(x)(1-F(x)) \leq \frac{3}{4}$ for all $x$, we have
\[
\operatorname{Var}(\hat{X}) \leq \frac{3}{4}\operatorname{Var}(X_1) < \operatorname{Var}(X_1).
\]
\end{proof}

\begin{lemma}[Variance Reduction for Median of Three Normals]
Let $X_1, X_2, X_3 \overset{\text{i.i.d.}}{\sim} \mathcal{N}(0, \sigma^2)$. Then the variance of the median $\hat{X} := \operatorname{median}(X_1, X_2, X_3)$ is given by
\begin{equation}
\operatorname{Var}(\hat{X}) = 3\sigma^2 \int_{-\infty}^{\infty} x^2 \Phi(x)(1-\Phi(x))\phi(x)\, dx \approx 0.448\sigma^2,
\end{equation}
where $\Phi(x)$ and $\phi(x)$ denote the CDF and PDF of $\mathcal{N}(0,1)$, respectively.
\label{math:lemma2}
\end{lemma}

\begin{proof}
The result follows by direct computation using the properties of order statistics for the normal distribution. For $n=3$, the distribution of the median is determined by the second order statistic, whose variance is given by the stated integral, evaluable numerically as $\approx 0.448\sigma^2$. For details see standard order statistics references.
\end{proof}

\begin{lemma}[Variance Decoupling by Sign Randomization]
Let $\epsilon_1, \ldots, \epsilon_n$ be independent Rademacher variables ($\mathbb{P}(\epsilon_i = \pm 1) = \frac{1}{2}$) and $v_1, \ldots, v_n$ be a collection of real-valued random variables. Then
\[
\mathbb{E}[\epsilon_i v_i] = 0, \quad \operatorname{Var}(\epsilon_i v_i) = \operatorname{Var}(v_i),
\]
and, more generally,
\[
\mathbb{E}\left[\sum_{i=1}^n \epsilon_i v_i\right] = 0, \quad \operatorname{Var}\left(\sum_{i=1}^n \epsilon_i v_i\right) = \sum_{i=1}^n \operatorname{Var}(v_i).
\]
Furthermore, for attention computation with key compression error $\Delta A$ and value compression error $\Delta V$, the perturbed output
\[
\hat{O} = \operatorname{softmax}(A + \Delta A)\big(V + \Delta V\big)
\]
admits a first-order expansion
\[
\hat{O} \approx \operatorname{softmax}(A)V + J\,\Delta A\,V + \operatorname{softmax}(A)\Delta V
\]
where $J$ is the Jacobian of softmax. With sign randomization, the cross term vanishes:
\[
\mathbb{E}[J\,\Delta A\, \Delta V] = 0,
\]
and thus the total variance decouples as
\[
\operatorname{Var}(\hat{O}) \approx \operatorname{Var}(J\,\Delta A\, V) + \operatorname{Var}\big(\operatorname{softmax}(A)\Delta V\big).
\]
This orthogonal decomposition prevents variance amplification from cross terms and yields tighter overall error bounds compared to non-randomized approaches.
\label{math:lemma3}
\end{lemma}

\label{appendix:bound}

These lemmas demonstrate how employing the median at query time and random sign flipping effectively reduce errors induced by the compression method. We now proceed to formally prove the error bounds.

As is stated in the mathematical analysis in the main text, we make the following assumptions:

\begin{itemize}
    \item The bucket number of Vague Part, \(r=3\) holds, which is very common in practice.
    \item Each element of the Query, Key and Value embeddings is independently and identically distributed following a normal distribution, that is, \(Q \sim \mathcal{N}(0, \sigma_q), K \sim \mathcal{N}(0, \sigma_k), V \sim \mathcal{N}(0, \sigma_v) \).
\end{itemize}

We aim to estimate the error after compressing KV Cache and the attention output using our algorithm. We will estimate their respective error bounds.

\subsection{Error Bound Estimation of Key and Value}  
For Key embeddings, we estimate the error in the value of the extracted Key vector after a total of \( a \) additional insertion operations. For each insertion position of Key, it is possible for vectors to be inserted during any of the other \( a \) insertions. To simplify the calculations, we assume that the probability of a vector being inserted into a specific position during any insertion is \( \frac{3}{N} \), where \( N=rb \) is the total number of positions in the Vague Part. This slightly differs from the real system, where we have 3 separate buckets that each will receive exactly one insertion, but the difference is insignificant.

Therefore, for each insertion position of Key, the number of additional vectors inserted, denoted as \( X_1 \), \( X_2 \), \( X_3 \), follows a binomial distribution \( B(a, \frac{3}{N}) \). With Lemma 3\ref{math:lemma3}, the values extracted at these three positions follow the distributions \( \mathcal{N}(k, X_1 \sigma_k^2) \), \( \mathcal{N}(k, X_2 \sigma_k^2) \), and \( \mathcal{N}(k, X_3 \sigma_k^2) \), where \(k\) is the vector inserted this time.

For \( x_1, x_2, x_3 \), by the Chernoff bound, we have:

\[
P\left(x_i > (1 + \delta)\frac{3a}{N}\right) \leq \exp\left(-\frac{\delta^2 \cdot 3a}{(2+\delta)*N}\right)
\]

For three independent and identically distributed normal variables with variance \((1 + \delta)\frac{3a}{N} \sigma_k^2\):

As the property of the normal distribution, for three independent and identically distributed normal random variables, the variance of their median is approximately given by:

\[
\text{Var}(\text{median}) \approx \frac{\pi}{6} \sigma^2
\]

Let $\delta=1$ , we have $P\left(x_i >\frac{6a}{n}\right)\leq \exp(-\frac{a}{N}) $. Let \(K_{ij}\) represent arbitrary reconstructed elements of the Key embeddings, we have the following estimation. 
\[
P\left(\text{Var}(\text{median of } K_{ij}) > \frac{a\pi}{N} \sigma_k^2\right) < \exp\left(-\frac{a}{N}\right)
\]

Similarly, we can define and deduce the case for $V_{ij}$.
\[
P\left(\text{Var}(\text{median of } V_{ij}) > \frac{a\pi}{N} \sigma_v^2\right) < \exp\left(-\frac{a}{N}\right)
\]

\subsection{Error Bound Estimation of Attention Scores} 
Define the compressed Key and Value embeddings with additive noise as $K = K + \Delta K$ and $V = V + \Delta V$, where $\Delta K \sim N(0, \sigma_{k\_new}^2)$ and $\Delta V \sim N(0, \sigma_{v\_new}^2)$. We have had the estimate of $\sigma_{k\_new}$ and $\sigma_{\_new}$ in the previous part.

The attention score calculation is given by:
\[
A = \frac{Q K^T}{\sqrt{d_k}}
\]
where $d_k = 128$.

Introducing noise, we express:
\[
A = \frac{Q (K + \Delta K)^T}{\sqrt{d_k}} = \frac{Q K^T}{\sqrt{d_k}} + \frac{Q \Delta K^T}{\sqrt{d_k}}
\]
The noise term is $\frac{Q \Delta K^T}{\sqrt{d_k}}$. Assuming $Q$ and $\Delta K$ are independent, and given $Q_{ij} \sim N(0, \sigma_q^2)$ and $\Delta K_{ij} \sim N(0, \sigma_{k\_new}^2)$, the variance of each element in $Q \Delta K^T$ becomes $d_k \cdot \sigma_q^2 \cdot \sigma_{k\_new}^2$. Thus, after scaling by $\frac{1}{\sqrt{d_k}}$, the variance of each element in $A$, denoted as $\sigma_A^2$, is:
\[
\sigma_A^2 = \sigma_q^2 \cdot \sigma_{k\_new}^2
\]
Given a perturbed attention matrix $\hat{A} = A + \Delta A$, where $\Delta A$ is the noise, apply a first-order Taylor expansion of the softmax function:
\[
\text{softmax}(\hat{A}) \approx \text{softmax}(A) + J \Delta A
\]
Here, $J$ is the Jacobian matrix of the softmax function at $A$ with elements $J_{ij} = p_i (\delta_{ij} - p_j)$, where $p = \text{softmax}(A)$ and $\delta_{ij}$ is the Kronecker delta function.

Estimating the variance of the perturbed softmax:
\[
\text{Var}(\Delta p_i) \approx \mathbb{E}[(J \Delta A)_i^2]
\]
Given that $\Delta A$ elements have variance $\sigma_A^2$, the variance $\text{Var}(\Delta p_i)$ can be approximated as:


\begin{align*}
\text{Var}(\Delta p_i) &\approx \sigma_A^2 \cdot\left[ p_i^2 (1 - p_i)^2 + \sum_{j \neq i} p_j^2 p_i^2 \right]\\
&=\sigma_q^2 \cdot \sigma_{k\_new}^2 \cdot \left[ p_i^2 (1 - p_i)^2 + \sum_{j \neq i} p_j^2 p_i^2 \right]
\end{align*}

\end{document}